\documentclass[sigconf]{acmart}
\acmSubmissionID{<PAPER ID>}
\usepackage{xspace}
\usepackage{subfig}
\usepackage{tabularx}
\usepackage{relsize}
\usepackage{enumitem}
\usepackage[linesnumbered, ruled, vlined]{algorithm2e}
\usepackage{algorithmicx}
\usepackage{algpseudocode}
\usepackage{tikz}
\newcommand*\circled[1]{\tikz[baseline=(char.base)]{
            \node[shape=circle,draw,inner sep=0.8pt] (char) {#1};}}

\RestyleAlgo{ruled}
\let\oldnl\nl% Store \nl in \oldnl 
\newcommand{\nonl}{\renewcommand{\nl}{\let\nl\oldnl}}% 
\newcommand\vldbdoi{XX.XX/XXX.XX}

\newcommand\vldbavailabilityurl{URL_TO_YOUR_ARTIFACTS}
\newcommand{\algo}{VT-GAN\xspace}
\newcommand{\tablegan}{{Table-GAN}\xspace}
\newcommand{\ctgan}{{CT-GAN}\xspace}
\newcommand{\ctabplus}{{CTAB-GAN+}\xspace}
\newcommand{\ctab}{{CTAB-GAN}\xspace}
\newcommand{\itgan}{{IT-GAN}\xspace}

\newif\ifignore
\begin{document}
% \title{\algo: Generating Tabular Data via Vertical Federated Learning}
\title{\algo: Cooperative Tabular Data Synthesis using Vertical Federated Learning}

%%
%% The "author" command and its associated commands are used to define the authors and their affiliations.
\author{Zilong Zhao$^*$}
\affiliation{%
  \institution{National University of Singapore}
  \streetaddress{P.O. Box 1212}
  \city{}
  \state{Singapore}
  \postcode{43017-6221}
}
\email{z.zhao@nus.edu.sg}

\author{Han Wu$^*$}
\orcid{0000-0002-1825-0097}
\affiliation{%
  \institution{University of Birmingham}
  \streetaddress{}
  \city{Birmingham}
  \country{UK}
}
\email{h.wu.6@bham.ac.uk}

\author{Aad van Moorsel}
\affiliation{%
  \institution{University of Birmingham}
  \streetaddress{}
  \city{Birmingham}
  \country{UK}
}
\email{a.vanmoorsel@bham.ac.uk}

\author{Lydia Y. Chen}
\orcid{0000-0002-1825-0097}
\affiliation{%
  \institution{TU Delft}
  % \streetaddress{P.O. Box 1212}
  \city{Delft}
  \state{The Netherlands}
}
\email{lydiaychen@ieee.org}

% \orcid{0000-0001-5109-3700}
% \affiliation{%
% \email{vb@rocquencourt.com}

% \author{J\"org von \"Arbach}
% \affiliation{%
%   \institution{University of T\"ubingen}
%   \city{T\"ubingen}
%   \country{Germany}
% }
% \email{jaerbach@uni-tuebingen.edu}
% \email{myprivate@email.com}
% \email{second@affiliation.mail}

% \author{Wang Xiu Ying}
% \author{Zhe Zuo}
% \affiliation{%
%   \institution{East China Normal University}
%   \city{Shanghai}
%   \country{China}
% }
% \email{firstname.lastname@ecnu.edu.cn}

% \author{Donald Fauntleroy Duck}
% \affiliation{%
%   \institution{Scientific Writing Academy}
%   \city{Duckburg}
%   \country{Calisota}
% }
% \affiliation{%
%   \institution{Donald's Second Affiliation}
%   \city{City}
%   \country{country}
% }
% \email{donald@swa.edu}

\begin{abstract}

%This paper presents the application of Vertical Federated Learning (VFL) to generate synthetic tabular data using Generative Adversarial Networks (GANs) while preserving privacy. VFL is a collaborative approach to training machine learning models among distinct tabular data holders, such as financial institutions, who possess disjoint features for the same group of customers. The authors present the VTS framework (Generating Tabular Data using VFL) and illustrate that VFL can be effectively used to implement GANs for distributed tabular data in a privacy-preserving way, achieving performance comparable to regular, centralized GANs that assume shared data. The paper outlines design choices in relation to the distribution of GAN generator and discriminator models and introduces a training-with-shuffling method to ensure that no participant can reconstruct the training data from the GAN conditional vector. An implementation of VTS and an in-depth evaluation of the quality and overall scalability of the VTS-generated synthetic data is presented for a range of data sets with diverse distribution characteristics. The findings demonstrate that VTS can consistently generate high-quality synthetic tabular data with a machine learning utility that differs by as little as 2.7\%, even in situations involving extremely imbalanced data distributions across clients or different numbers of clients.

This paper presents the application of Vertical Federated Learning (VFL) to generate synthetic tabular data using Generative Adversarial Networks (GANs). VFL is a collaborative approach to train machine learning models among distinct tabular data holders, such as financial institutions, who possess disjoint features for the same group of customers. In this paper we introduce the \algo framework, Vertical federated Tabular GAN, and demonstrate that VFL can be successfully used to implement GANs for distributed tabular data in privacy-preserving manner, with performance close to centralized GANs that assume shared data. We make design choices with respect to the distribution of GAN generator and discriminator models and  introduce a training-with-shuffling technique so that no party can reconstruct training data from the GAN conditional vector. The paper presents (1) an implementation of \algo, (2) a detailed quality evaluation of the \algo-generated synthetic data, (3) an overall scalability examination of \algo framework, (4) 
%a membership inference attack on \algo with differential privacy protection
a security analysis on \algo's robustness against Membership Inference Attack with different settings of Differential Privacy, 
for a range of datasets with diverse distribution characteristics. Our results demonstrate that \algo can consistently generate high-fidelity synthetic tabular data of comparable quality to that generated by a centralized GAN algorithm. The difference in machine learning utility can be as low as 2.7\%, even under extremely imbalanced data distributions across clients or with different numbers of clients. 

% following are the abstract used for first GTV version
% This paper introduces the use of Vertical Federated Learning (VFL) for privacy-preserving generation of synthetic tabular data using Generative Adversarial Networks (GANs). VFL is a natural approach to collaboratively training machine learning models among different tabular data holders such as financial companies that hold disjoint features for the same set of customers. In this paper we introduce the \algo framework, Generating Tabular Data using VFL, and demonstrate that VFL can be successfully used to implement GANs for distributed tabular data in privacy-preserving manner, with performance close to regular, centralised, GANs that assume shared data. We make design choices with respect to the distribution of GAN generator and discriminator models, and we introduce a training-with-shuffling technique so that no party can reconstruct training data from the GAN conditional vector. The paper presents an implementation of \algo and a detailed evaluation of the quality and overall scalability of the \algo-generated synthetic data, for a range of data sets with diverse distribution characteristics. Our results demonstrate that \algo can consistently generate high-fidelity synthetic tabular data of comparable quality to that generated by a centralized GAN algorithm. The difference in machine learning utility can be as low as 2.7\%, even under extremely imbalanced data distributions across clients or with different numbers of clients. 

\ifignore
Generative Adversarial Networks (GANs) have achieved state-of-the-art results in tabular data synthesis under the presumption of directly accessible training data.
%that training data is centrally available. 
%However, due to data privacy concerns, this assumption is not always tenable. 
% Generative Adversarial Networks (GANs) have shown state-of-the-art results in synthesizing tabular data, but this requires central access to the training data, which is not always possible due to privacy concerns. 
Vertical Federated Learning (VFL) is a paradigm which allows to distributedly train machine learning model with clients possessing unique features pertaining to the same individuals, where the tabular data learning is the primary use case.  %. Due to its nature, tabular data is the primary focus for VFL applications.
%Given the unique nature of tabular data, in a distributed system where different clients possess unique features pertaining to the same individuals, sharing data is not allowed by privacy laws.
%(e.g., General Data Protection Regulation (GDPR) in EU). 
%Sharing this data would have mutual benefits for building better machine learning models for all parties, but it is often prohibited by laws (e.g., General Data Protection Regulation (GDPR) in EU).
% The paradigm of Vertical Federated Learning (VFL) presents a viable solution for the above scenario.
%for the distributed training of machine learning models, where different clients possess distinct data features.
However, 
it is unknown if tabular GANs can be learned in VFL. Demand for secure data transfer among clients and GAN during training and data synthesizing poses extra challenge.
% there is no prior research exploring the integration of tabular GANs into VFL. 
% GAN consists of two networks generator and discriminator, securely coordinating the data transfer among clients and GAN requires careful design. 
Conditional vector for tabular GANs is a valuable tool to control specific features of generated data. 
But it contains sensitive information from real data - risking privacy guarantees.
%But the inclusion of sensitive information from real data within the conditional vector poses a challenge for its implementation within VFL.
% can be used to control specific features of the generated data.
%Additionally, the feasibility of incorporating the conditional vector into above training without privacy leakage remains uncertain. 
In this paper, we propose \algo, a VFL framework for tabular GANs, whose key components are generator, discriminator and the conditional vector.
%for generating synthetic tabular data using GANs via VFL. 
% The system is designed to contain a trusted third-party server and multiple clients. 
\algo proposes an unique distributed training architecture for generator and discriminator to access training data in a privacy-preserving manner.
To accommodate conditional vector into training without privacy leakage, \algo designs a mechanism \textit{training-with-shuffling} to ensure that no party can reconstruct training data with conditional vector.
% GTV also adopts the conditional vector into training without compromising privacy by incorporating a mechanism called \textit{training-with-shuffling} 
% % . All the clients shuffle their local data in the end of each training round 
% so that neither the server and the clients can infer sensitive information from the system. 
% We have extensively evaluated GTV using five datasets and eight metrics, and the 
We evaluate the effectiveness of \algo in terms of synthetic data quality, and overall training scalability. 
Results show that \algo can consistently generate high-fidelity synthetic tabular data of comparable quality to that generated by centralized GAN algorithm. The difference on machine learning utility can be as low as to 2.7\%, even under extremely imbalanced data distributions across clients and different number of clients. 
%\lc{Give a concrete number about the improvement}
% Furthermore, GTV maintains stable for synthetic data column distributions with an increasing number of clients, 
% making it a robust and scalable solution for generating synthetic tabular data in a distributed and privacy-preserving manner.
\fi
\end{abstract}
\maketitle
\def\thefootnote{*}\footnotetext{Equal contribution}
\section{Introduction}
\label{sec:intro}
% \lc{I found the first three paragraph too lengthy. Instead, I think the fourth paragraph shall be extended - to illustrate the difficulties and challenges of VFL for GAN }
% Modern technology companies rely on data analytics to enable automation, improve decision-making and personalize services. Meanwhile, emerging data protection rules such as the General Data Protection Regulation (GDPR) enforce strict constraints on the use of personal data. A solution is to leverage synthetic data from machine learning models.
%Particularly, companies cannot directly share their data that contain personal or sensitive information. A solution is leveraging synthetic data. 
% In this context, data synthesis technology is developed to provide both robust privacy protection and the possibility to generate useful data when real data is not accessible.
% \lc{This pargraph should be about tabular GAN already}\zz{difficult to introduce tabular data here}Generative adversarial networks (GANs)~\cite{gan} are a powerful class of machine learning models that are capable of generating synthetic data that is similar to real data. GANs have been successfully applied to tasks such as image synthesis~\cite{stylegan3} and style transfer~\cite{cyclegan}. 
% Given that tabular data, which is organized in rows and columns, is the most prevalent data format in industry~\cite{arik2019tabnet}, there is growing interest in developing GANs for table synthesis. 
Tabular data, organized in rows and columns, is the most prevalent data format in industry~\cite{arik2019tabnet}. Recent years has seen the proliferation of the use of synthetically generated tabular data as a privacy-preserving approach for data analysis and product development~\cite{medgan,mottini2018airline}.  State-of-the-art (SOTA) generation of synthetic tabular data utilises Generative Adversarial Networks (GANs)~\cite{gan,ctgan,tablegan,ctabgan,ctabplus,itgan}.  
However, training these GANs still requires direct access to all training data, raising an additional privacy concern if the data is coming from multiple sources. 

Consider the following scenario. A regional e-commerce company and a bank hold separate information for a set of shared customers. Generating a synthetic dataset that combines the bank's customer income records with the customer's purchasing history in the e-commerce company would be highly beneficial to create more valuable data analysis and associated products. For instance, the bank can build a more robust prediction model for credit rating and the e-commerce company can build a more accurate recommendation system for its customers. In such cases, collecting and combining data from the bank and e-commerce company to train tabular GANs is not an option, because it involves sharing personal customer information across different organizations.  In other words, we require a privacy-preserving approach to training the GAN, which avoids sharing data, and in this paper we consider Vertical Federated Learning (VFL). 
%Without centralized access to the entire dataset, 
%Given this, it is crucial to develop a distributed and privacy-preserving 
% \lc{privacy-preserving? Use one of them thoroughout the paper }
%solution that can generate a comprehensive synthetic dataset without compromising customers' personal information. 

%In response to such demand, vertical federated learning (VFL) paradigm emerges. 
Federated learning is a widely explored technique to train machine learning models on distributed data without sharing the data. Data remains in participating parties that hold data (known as clients) and learning is achieved by exchanging model information with a central party (also called the server), without sharing the data itself. In {\em Vertical} Federated Learning clients hold a local dataset with unique features pertaining to the {\em same} individuals or other identified subjects. This makes VFL particularly appropriate for various tabular data applications, while image or audio data would more naturally utilise {\em horizontal} federated learning, in which all clients have data with the same data feature structures. To train a prediction model under VFL~\cite{wei2022vertical}, each client passes its local data through a bottom model (on the client side) and sends the output to a top model (on the server side) to calculate gradients. This ensures that local data is not directly accessed by any party except the owning client, and only intermediate results are shared. 
%The final VFL model is a combination of the top and bottom models. 
Previous studies of VFL have concentrated on prediction models such as decision trees~\cite{wu2020privacy} and deep neural networks~\cite{romanini2021pyvertical}, 
%{but the use of generative models is also important.} 
%Training prediction model in VFL requires a pre-defined label, and clients are unable to independently conduct data cleaning or feature engineering on all data. By contrast, after training a generative model (e.g., GANs), clients can conduct 
% The knowledge that we can extract from the synthetic data generated from all clients' data is far beyond the construction of a prediction model. Synthetic data can facilitate the identification of more useful information and even lead to more collaborations between organizations. 
% The insights that can be obtained from synthetic data generated from all clients' data surpasses the meaning of a prediction model. The utilization of synthetic data allows for a more thorough understanding of the data, leading to the identification of valuable information and potential opportunities for more collaboration among organizations.
% Despite this potential,
there is no platform to train tabular GANs using VFL.
%there is no research on training generative model, such as GANs, for tabular data synthesis under the VFL paradigm.
%VFL is a distributed machine learning framework that enables organizations to collaboratively train machine learning models on decentralized data without requiring the raw data to be shared. In VFL, each client holds a local dataset with its own unique features pertaining to the same individuals, this character makes it a perfect fit for tabular data instead of image, audio or text. During training, each client passes local data through a bottom model (in client size), then sending the output to top model (in server side) to calculate gradients. 
%The model in VFL often consists of two parts: (1) top model in the server (usually located in the client who holds the labels) and (2) bottom model in each client. Clients pass local training data through bottom model and share the intermediate results with server. Server concatenates the intermediate results and use it as the input for top model. Top model will calculate loss and back-propagates the gradients in the backward direction from the top model to all the bottom models. 
% As such, the local data is not directly accessed, except by the owner, and only intermediate result is shared. Eventually, the top and bottom models together is the final VFL model. Previous studies of VFL are mostly based on prediction models, there is no prior research on training GANs for tabular data under the VFL paradigm.

\ifignore
Training of tabular GANs, e.g., \ctgan~\cite{ctgan} and CTAB-GAN~\cite{ctabgan}, for VFL present multiple challenges. 
% how can the GAN model learn cross-client data dependency?
Firstly, one needs to assure that the synthetic data does not only retain data correlations for the data within one client, but also captures the column dependencies across the data in different clients.
% how to incoporate CV into GAN via VFL in a privacy-preserving manner?
Secondly, many advanced tabular GANs utilize a technique called conditional GAN (CGAN). Assuming there are two types of columns in tabular data: categorical and continuous. The inclusion of a conditional vector (CV)\cite{ctgan} allows for control over the class of generation in certain categorical columns (e.g., indicating 'male' or 'female' in \textit{gender} column). This is particularly useful when the class distribution is imbalanced in that column, because after constructing a CV, a corresponding row from the real dataset must also be selected for training data based on the class indicated in the CV. By carefully designing the CV, it is possible to balance the class representations in the training data.
% it is possible to over-sample minority categories in certain columns to rebalance the distribution of the training data. 
% In the proposed VFL method, 
% the CV must be able to indicate the categories in all categorical columns from all clients. 
But with the constructed CV, 
% a part of real data is also selected as training data. 
the indices of the selected training data are available to clients or server. For any party who has the access to both the data indices and CVs, it can potentially reconstruct categorical columns of training dataset from all clients.
%This presents a privacy concern as both clients and the server can potentially reconstruct information within the categorical columns of training dataset from all clients by having access to both the data indices and CVs. 
Therefore, a privacy-preserving mechanism is needed to accommodate the CV in the \algo training process. 
Thirdly, GAN consists of two parts: generator and discriminator, only the discriminator needs to access real data. 
There is no prior knowledge on whether we should partition both the generator and discriminator into top and bottom models 
% \lc{top and bottom models are not introduced yet}\zz{in line 20} 
since the generator is already isolated from real data. If so, we need to find the optimal partitions so that the final generator can produce high-fidelity synthetic data.
\fi 

{In this paper, we present a novel framework -- \textbf{\algo} 
which implements \textbf{V}ertical federated learning for \textbf{T}abular \textbf{GAN}.}
% which provides an infrastructure to train GANs for \textbf{G}enerating \textbf{T}abular data via \textbf{V}ertical federated learning.  
The main challenge is to determine the bottom and top models, such that no data or personal information leaks from the individual clients, and the performance is close to that of a centralized model. 
%For synthetic data generation, GANs are the model of choice, and 
By partitioning both GAN generator and discriminator models, placing pieces in each client and in the server, we assure that no data needs to be exchanged but that correlations between data from different clients can be captured by passing intermediate outcomes to the top model. 
%A main contribution of this paper is the extensive evaluation of different partitions of state-of-the-art tabular GANs, depending on the distribution of data across clients. It demonstrates the viability of VFL for synthetic tabular data generation and also provides insights in how to partition for varying scenarios. 
Advanced tabular GANs utilize a technique called conditional GAN (CGAN)\cite{ctgan} and \algo supports CGANs.  Conditional GANs use a {\em conditional vector} (CV) to express conditions about the data considered in each learning step (e.g., limited to a specific gender). In \algo, the server sends the conditional vector to all clients and the clients select data for their bottom model accordingly. 
However, a challenge arises when facilitating CGANs: the server may potentially deduce data characteristics in the clients based on the responses they receive from their bottom models, which are subsequently shared with the server.
% There is a challenge in facilitating CGANs in that the server may be able to infer characteristics about the data in the clients based on the responses that the clients obtain from their bottom model and which they send to the server. 
To mitigate this risk, in the end of each training round, \algo implements a mechanism called \textit{training-with-shuffling} in which all clients shuffle their local data using the same random seed. This ensures that the data after shuffling is still consistent across clients, but is not known to the server. Through this mechanism, the mapping between the CV and data index changes each round, making it impossible for the server to reconstruct the training data of clients.

% Since we consider synthetic data generation, our approach differs somewhat from the traditional application of VFL for prediction or classification models depicted in Fig.\ref{fig:vfl_classifier}. In particular, we establish a separate server that maintains the top model, instead of appointing as server the client that contains data associated with a specific label. We assume the server to be honest but curious, as are all the clients. We note that most of the VFL literature assumes the honest-but-curious threat model \cite{wu2020privacy, fang2021large, fu2021vf2boost, fu2022blindfl}, implying that clients and server correctly execute the protocol, but do seek to gain information about data in other clients while running the protocol. 

\ifignore
% \lc{We are missing the privacy challenge of dealing with conditions in conditional GAN}
In this paper, we  
%a novel framework  a set of rules and guidelines that provides structure for \textbf{G}enerating \textbf{T}abular data in \textbf{V}ertical federated learning - GTV.
present a novel framework -- \textbf{\algo} which provides an infrastructure to train GANs for \textbf{G}enerating \textbf{T}abular data via \textbf{V}ertical federated learning.  \algo partitions both the generator and the discriminator into top and bottom models to ensure the server cannot infer sensitive information from clients' data. 
% The optimal partitions are discussed under different data distribution and varying number of clients. 
Unlike the traditional VFL setting for building prediction models which places the server in the client who holds the label column of tabular data, 
% To build prediction models in VFL, the training data must contains one label column, and the server is placed in the client (i.e., label holder) who holds the label column. 
\algo places the server in a trusted third-party. The reasons are: (1) for training GANs (i.e., a type of generative model), it is not necessary to have a label column in the training data and (2) since the server always has knowledge of the CV, if the server is placed in one of the clients, that client would also have knowledge of the data index for each CV, making it possible to reconstruct data columns of other clients.
%\algo holds the server in trusted third-party instead of in the client who holds the labels, which is the traditional VFL setting for building prediction model, because for tabular data synthesis, it is not necessary to contain a label column in the table. It is also because server always has the knowledge of conditional vector. If the server resides in the label holder, it will also have the knowledge of data index for each condition. That will make the label holder possible to reconstruct data columns of other clients.
% However, even server does not reside in the label holder, if the data index remains the same for all the training rounds, the server can still reconstruct the training data from all the clients using conditional vector and its corresponding data index. Therefore, in the end of each training round, all the clients will shuffle their training data using the same random state so that the data after shuffling are still corresponding across clients. And server should be isolated from the shuffle function and random state that clients use. We name above training mechanism \textit{training-with-shuffling}. 
However, even if the server does not reside in the client, 
%if the data indices remain the same for all the training rounds, 
it can still reconstruct the categorical columns from training data using the CVs and its corresponding data indices. {Data index and CV are like a pair of coordinate, they can indicate a presence of a specific category in a categorical column for a given data index row.} 
To mitigate this risk, in the end of each training round, \algo implements a mechanism called \textit{training-with-shuffling} in which all clients shuffle their local data using the same random seed. This ensures that the data after shuffling is still consistent across clients.
The server is isolated from the shuffle function used by clients. This mechanism ensures that the mapping between CV and data index changes each round, making it impossible for server to reconstruct training data of clients.
% server cannot reconstruct the training data from all the clients using the CV and its corresponding data index because index changes every training round.
\fi 

\begin{figure}[t]
    \centering
    \includegraphics[width=0.80\linewidth]{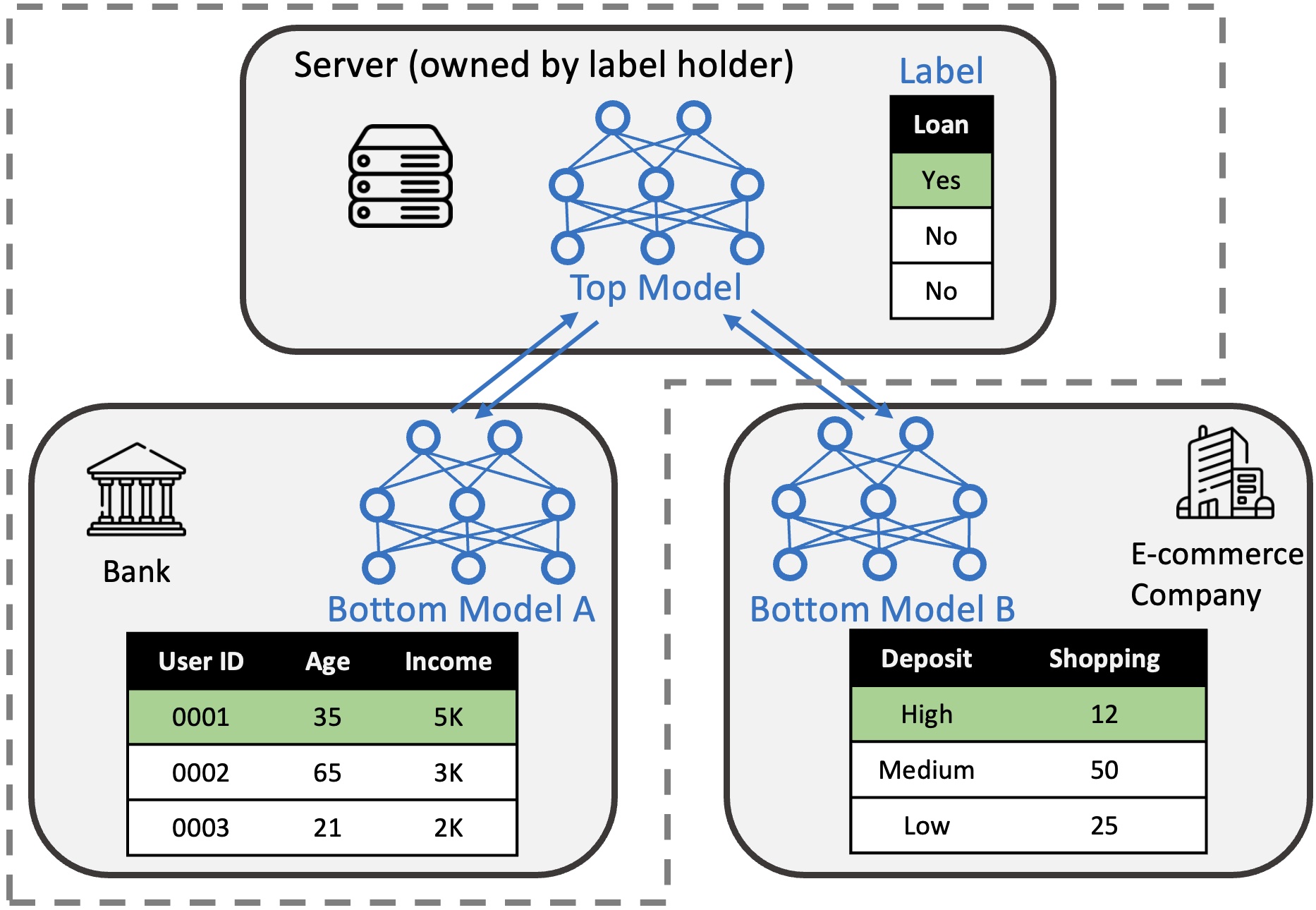}
    \vspace{-0.5em}
    \caption{Traditional VFL architecture for prediction model.}
    \label{fig:vfl_classifier}
   \vspace{-1em}
\end{figure}

% We extensively evaluate the \algo with nine partitions of the generator and the discriminator neural networks, and compared with centralized tabular GAN as baseline. We also evaluate the \algo with different data distribution across the clients. The experiments are conducted on five commonly used machine learning tabular datasets where the machine learning utility difference and statistical similarity between real and synthetic data are reported. Results show that as long as the top model of discriminator is big enough, the \algo can perform closely as centralized tabular GAN. And for data distribution experiment, we found that the less the features in label holder, the worse the machine learning utility of the synthetic data. And by making the top model of the generator and discriminator as big as possible, \algo can reduce the impact of the imbalanced data distribution. 
We extensively evaluate \algo using nine different partitions of the generator and discriminator neural networks between server and clients, and compare it to a centralized tabular GAN. 
% We also evaluate \algo under different data distribution scenarios across the clients. 
The experiments are conducted on five commonly used tabular datasets in machine learning.
%, all of which contain target columns. 
We report on the machine learning (ML) utility and statistical similarity difference between the real and synthetic data. Results show that as long as the top model of the discriminator is large enough, \algo performs comparably to the centralized tabular GAN baseline. In the data partition experiments, we discover that the more imbalanced the number of data columns across the clients, the lower the ML utility of the synthetic data. However, by shifting more neural networks of the generator from bottom to top model, \algo mitigates the negative effects of this imbalance. Additionally, we find that with an increasing number of clients, expanding the size of the generator neural network helps counteract the degradation in synthetic data quality. {Finally, we evaluate the vulnerability of \algo to Membership Inference Attacks (MIA). The full black-box MIA from \cite{chen2020gan} on \algo generated synthetic data shows low success rate. Implementing Differential Privacy (DP) on \algo can further lower the success rate of MIA, albeit at the expense of \algo's performance.}
%In the experiment with increasing number of clients, enlarging neural network size of generator help counter the degradation of synthetic data quality.
The main contribution of this study can be summarized  as follows:
% \zz{add code link somewhere.}
% \zz{maybe training time discussion?}

\begin{itemize}[leftmargin=*]
\item We design the first of its kind distributed framework \algo which incorporates state-of-the-art tabular GANs into vertical federated learning architecture. 
%Code is provided here\footnote{\url{https://drive.google.com/file/d/1J0-7GNASpDOn0F8O1l0Hw_9mAQvrmM-j/view?usp=sharing}}.
%Results show that the \algo is able to successfully generate synthetic data that captures column dependencies of the data across the clients in different locations.

% \item 
% To address the challenge of adopting conditional vectors in GTV without privacy loss, we design the \textit{training-with-shuffling} mechanism to prevent both server and clients from inferring sensitive information. 
\item To accommodate conditional vector into \algo training in a privacy-preserving manner, we devise the \textit{training-with-shuffling} mechanism.
%We devise the \textit{training-with-shuffling} mechanism to accommodate conditional vector into GTV training in a privacy-preserving manner.
%We design a mechanism \textit{training-with-shuffling} to incorporate conditional vectors in GTV 
We also propose a secure synthetic data publication strategy to mitigate the risk of inference attacks.

\item  {We consider the semi-honest threat model and use it to motivate the design of \algo. We qualitatively analyze the potential privacy risks in \algo training. We experimentally evaluate the robustness of \algo against Membership Inference Attack.
%We also quantitatively examine the potential privacy leakage by the synthetic data using Membership Inference Attack. 
Differential Privacy is also studied to implement on \algo to provide protection.
} 
%We analyze possible privacy issues during the training and data generation process in \algo, strictly define the information that each party can obtain, and elaborate the defensive capabilities of \algo against commonly studied attacks\lc{Write out the attacks}. \lc{After reading the entire paper, I will suggest to take out this argument. This will lead to many attacks from the "security" reviewers}
% \lc{where?}.
% \lc{Suggestion: 1. We qualitatively analyze the potential privacy risk. 2. We first talk about the threat models and use that motivate the design of GTV}

\item We extensively evaluate \algo using five widely used machine learning datasets on eight evaluation metrics including the aspects of machine learning utility and statistical similarity. Results show that \algo can stably generate high-fidelity synthetic data under extremely imbalanced data distribution across the clients. Optimal setup for \algo is recommended under the constraints of computation resources, budgets, number of clients and data distributions across clients.
% \lc{I thought we discussed to make it as our algorithm to decide the optimal split}

% \item \lc{highlight the results}
\end{itemize}

% Synthetic data which contains features from both the e-commerce company and the bank is beneficial, but we need a decentralized privacy-preserving solution to approach that.

% \textbf{Background and motivation} background of a) synthetic data b) Vertical Federated Learning. Include the figure of five datasets' model performance in the motivation.
% \begin{figure}[htb]
%     \centering
%     \includegraphics[width=0.99\linewidth]{figures/motivation.png}
%     %\vspace{-1em}
%     \caption{Motivation Case Studies}
%     \label{fig:motivation}
%    %\vspace{-1em}
% \end{figure}
% \textbf{Challenges}
% \zz{put the motivation case study figures and discussion here}

% \textbf{Summary of Contributions}

\vspace{-0.5em}
\section{Preliminaries and Motivation}
% \zz{introduce basic VFL and GAN training flow. } plot a workflow figure of VFL, client A and Client B (label holder).
% \wh{introduce tabular data structure, especially the difference from unstructured data (images, text)}
\algo enables training SOTA tabular GANs on VFL. In this section, we describe the preliminary of VFL and GAN.
\vspace{-0.2em}
\subsection{Vertical Federated Learning}
As depicted in Fig.\ref{fig:vfl_classifier}, a typical VFL system involves multiple clients, such as a bank and an e-commerce company, which possess distinct features for a shared group of users. It is important to note that in the illustration, the data instances of the clients have been aligned such that each row corresponds to the same individual across all clients. % However, in real-world scenarios, it is possible for clients to have instances that are not present in other clients' data. 
{Data alignment across clients can be achieved by the Private Set Intersection (PSI) technique \cite{dong2013private,chen2017fast}, which is a common assumption used by VFL studies \cite{wu2020privacy,fu2021vf2boost,fu2022blindfl}. We also adopt it into our \algo.}
% without further explanation in the paper. 
% Therefore, it is out of scope of the discussion in this paper.
% In this study, we do not address this issue and assume that data alignment can be addressed through the use of the Private Set Intersection technique \cite{dong2013private,chen2017fast}, a common approach in other VFL studies \cite{wu2020privacy,fu2021vf2boost,fu2022blindfl}. 
% In real-world scenarios, it is possible for clients to possess overlapping features. In this study, we assume that clients already resolve any overlap by excluding redundant features, resulting in a system where all clients possess only unique features.

The goal of VFL is to train a common machine learning model using the features from all clients. 
%Thus the input of a VFL task is the virtually joined dataset from the clients. 
In VFL, there are two types of models: top and bottom models as represented 
%, which are represented by the fully connected neural network blocks 
in Fig.\ref{fig:vfl_classifier}. 
% In VFL, there are top and bottom models, represented by the fully connected neural network blocks in Fig.\ref{fig:vfl_classifier}. 
During the training, each client passes their local data through the bottom model and sends the intermediate logits to the server. In practice, the server is typically managed by the client that holds the label column. As depicted by the dotted line in Fig.\ref{fig:vfl_classifier}, Bank is the label holder. The server horizontally concatenates the intermediate logits from all the clients and uses the concatenation as input for the top model. 
The top and bottom models are then updated using the gradients calculated from the discrepancy between the output of the top model and the label.
% The discrepancy between the output from the top model and the label is then used to calculate the gradients for updating the top and bottom models.
\begin{figure}[t]
    \centering
    \includegraphics[width=0.80\linewidth]{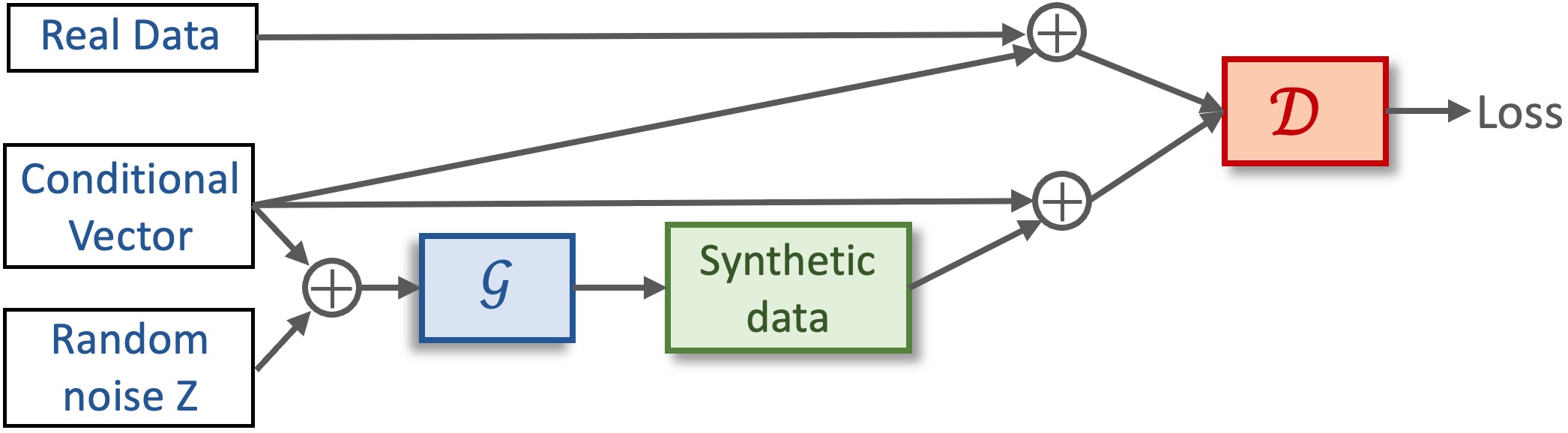}
    \vspace{-0.5em}
    \caption{Conditional GAN (centralized).}
    \label{fig:conditional_GAN}
   \vspace{-1.8em}
\end{figure}
 
% \vspace{-0.5em}
\vspace{-0.5em}
\subsection{Tabular Generative Adversarial Network} 
%\lc{I think the emphaseis to be one the tabular GAN?} \lc{we can change this to Conditional GANs}
Generative Adversarial Networks (GANs)~\cite{gan} are a type of machine learning model designed to generate synthetic data that is similar to a given training dataset. They are composed of two parts: a generator ($\mathcal{G}$) and a discriminator  ($\mathcal{D}$) as illustrated in Fig.~\ref{fig:conditional_GAN}. The generator is trained to generate synthetic data that is similar to the real data, while the discriminator is trained to distinguish between the synthetic data generated by the generator and the real training data. 
During the $\mathcal{D}$ training step, the discriminator is trained separately by the synthetic data from generator and real data. During the $\mathcal{G}$ training step, only the synthetic data is used. 
%The discriminator evaluates the quality of the synthetic data and provides the loss to update the generator. 
% The generator and discriminator play a two-player minimax game, where the generator tries to produce synthetic data that is indistinguishable from the real data, and the discriminator tries to accurately classify the synthetic data. 
This process continues until the generator produces synthetic data that is indistinguishable from the real data.
%, at which point the GAN has succeeded in learning the distribution of the real data.

% During training, the generator and discriminator play a two-player minimax game, where the generator tries to produce synthetic data that is indistinguishable from the real data, and the discriminator tries to accurately classify the synthetic data as fake. This process continues until the generator produces synthetic data that is indistinguishable from the real data, at which point the GAN has succeeded in learning the distribution of the training data.

To better indicate the category of generation (e.g.,  'male' or 'female' in \textit{gender} column), conditional GAN is well adopted for table synthesis~\cite{ctgan, ctabgan, ctabplus}. Fig.~\ref{fig:conditional_GAN} shows the structure of a conditional GAN (CGAN). CGAN controls the generation of synthetic data 
%allows the generation of synthetic data to be controlled 
through the use of an auxiliary conditional vector (CV). The generator and discriminator are conditioned on the same CV. When using the real data to train discriminator, once a CV is given, the real data needs to sample one row of training data whose class is corresponding to the condition indicated in the CV. 
%\lc{Give a concrete example of using condition vector}
%This can be useful when the real data is imbalanced, as the CGAN can be used to class representations in training data.
Feature engineering is another important tool for training tabular GANs. Before entering real data to $\mathcal{D}$, \cite{ctgan, ctabgan, ctabplus, itgan} use mode-specific normalization~\cite{ctgan} to encode continuous columns, one-hot encoding for categorical columns, mixed-type encoding~\cite{ctabgan}
for column contains both categorical and continuous values.

\vspace{-0.1em}
\section{Methodology}
In this section, we begin by providing an overview of \algo. We then justify the design of \algo by introducing the privacy and security consideration.
%\lc{many different superscripts are used. The conventions need to explained right at the beigning}

\begin{table}[t]
\small
\centering
\caption{Summary of Notations}
\vspace{-1.5em}
\begin{tabular}{ll}
\toprule
\textbf{Notation} & \textbf{Description}                                 \\ \hline
$\mathcal{G}^t$, $\mathcal{D}^t$ & top models of generator, discriminator  \\
$\mathcal{D}^s$ & conditional vector filter which runs on server \\
$\mathcal{G}^b_i$, $\mathcal{D}^b_i$ & bottom model of generator, discriminator on client $i$\\
$T_{i}$ & real tabular data held by client $i$\\
$CV_p$ & conditional vectors constructed by client $p$\\
$idx_p$ & data indices, the corresponding data contains\\
& the indicating classes in $CV_p$\\
$P_r$ & ratio vector, the ratio of number of features in each \\ & client to the total number of features across all the clients\\
\bottomrule
\end{tabular}
\label{tab:symbol}
\vspace{-1.5em}
\end{table}

\vspace{-0.1em}
\subsection{\algo Architecture}
In this part, we begin by discussing the components of \algo with an explanation of the design rationale for the structure, then introduce the threat model of our framework followed by a walkthrough of the training process. We then delve into the specifics of training, including possible adjustments. Secure design for publishing synthetic data is explained in the end.
% We will first introduce the components of \algo, then go through the training flow, and then focus on the training details. The rationality of the structure design and potential adjustment is provided as an extended discuss.
% We will first explain the general training flow, then explain the components of \algo and the rationality of structure design. In the end, we will focus on the training details.
%\lc{Introduce the components first before the training flow}
%\lc{My proposal to the subsections: 4.1 GTV architecture, 4.2 GTV training procedure, 4,2,1 training by shuffling}

% \subsubsection{Threat Model}
% In line with many previous works in the field of VFL \cite{wu2020privacy, fang2021large, fu2021vf2boost, fu2022blindfl}, we focus on the semi-honest model in the design of our proposed GTV architecture. Specifically, we assume that the clients and server are honest but curious, meaning that they adhere to the GTV protocol, but may seek to acquire additional information through the computation process. We do not consider collusion between server and clients. Despite the fact that tabular datasets are retained locally within each client, the values derived during the GTV learning and synthesizing process may also reveal sensitive information and result in data leakage. 

\subsubsection{\algo overview}
\begin{figure}[t]
    \centering
    \includegraphics[width=0.99\linewidth]{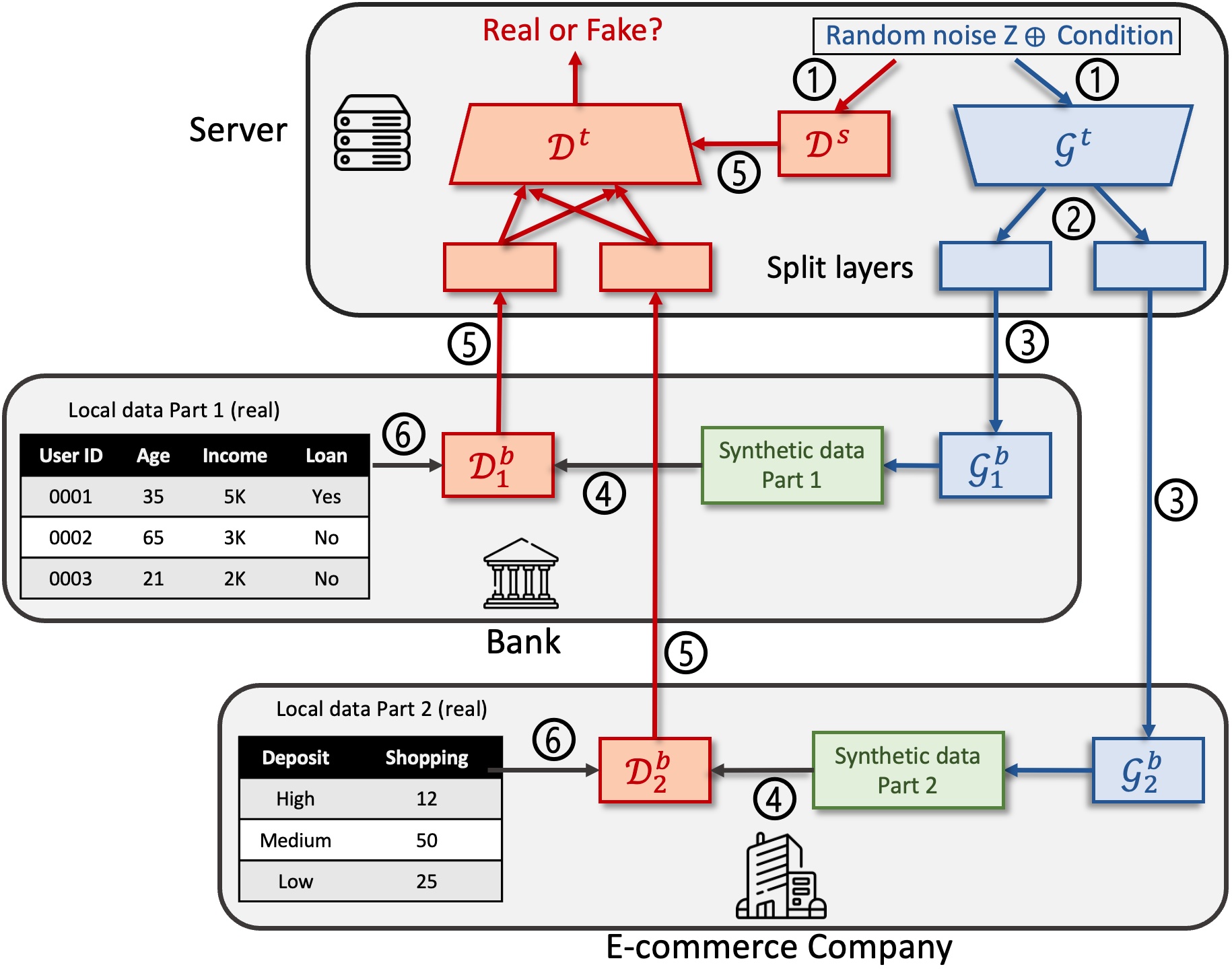}
    %\vspace{-1em}
    \vspace{-0.5em}
    \caption{The workflow of \algo.}
    \label{fig:gtv_workflow}
   \vspace{-1em}
\end{figure}

% \wh{add numbers to the arrows in \algo figure}
Fig.~\ref{fig:gtv_workflow} shows an example of \algo with two clients. The notions of the components are provided in Tab.~\ref{tab:symbol}. Each client contains unique data columns. 
\algo establishes a separate server instead of assigning the client that holds the label as server, as in Fig.~\ref{fig:vfl_classifier}.
% Unlike traditional VFL, \algo does not reside server on one of the clients (i.e., the label holder for training prediction model in VFL), 
% but hosts by a trusted third-party.
% The reason is (1) for table synthesis, the tabular data does not necessarily contain a target column, therefore, there is no label holder, (2) due to the privacy concern, we do not want server holder have access to the \texttt{shuffle()} function that client uses to shuffle local data, we will detail this reason when we explain the  \texttt{shuffle()} function in Sec.~\ref{sssec:training_details}. In fig.~\ref{fig:gtv_workflow}, another noticeable part is that we intentionally split part of the discriminator and the generator from server side to client side, i.e., $\mathcal{D}^b_1$, $\mathcal{D}^b_2$, $\mathcal{G}^b_1$, $\mathcal{G}^b_2$. This design is also due to privacy concerns. First of all, $\mathcal{D}^b_1$ and $\mathcal{D}^b_2$ exist because clients cannot send their real data directly to server. And $\mathcal{G}^b_1$ and $\mathcal{G}^b_2$ are needed because otherwise the output from $\mathcal{D}^t$ will only go through $\mathcal{D}^b_1$, $\mathcal{D}^b_2$ and goes back to server. Then the server can reverse-engineer the $\mathcal{D}^b_1$, $\mathcal{D}^b_2$, then reconstruct clients' local data. 
The reason is twofold: (1) In table synthesis, when label column is not needed, no label holder exists. (2) For privacy reasons, clients should not access both the CVs and selected data indices. The server also should not  have access to the \texttt{shuffle} function that the client uses to shuffle local data (we discuss this function in more detail in Section~\ref{sssec:training_details}). 
% \lc{Though the notations defined differently from the evaluation, but they are similar. So, it is misleading to the reviewers what is what}
In Fig.~\ref{fig:gtv_workflow}, one notable feature is the partition of the discriminator and generator between the server and the clients. The design of the bottom models, i.e., $\mathcal{D}^b_1$,  $\mathcal{D}^b_2$ and $\mathcal{G}^b_1$, $\mathcal{G}^b_2$, are also motivated by privacy concerns. First, $\mathcal{D}^b_1$ and $\mathcal{D}^b_2$ are used because clients cannot directly send their real data to the server. Second, $\mathcal{G}^b_1$ and $\mathcal{G}^b_2$ are necessary because otherwise, the output from $\mathcal{G}^t$ would only go through $\mathcal{D}^b_1$, $\mathcal{D}^b_2$ and return to the server. If so, the server could reverse-engineer $\mathcal{D}^b_1$, $\mathcal{D}^b_2$ and reconstruct the clients' data.
% One may also doubt the rationality of the existence of $\mathcal{D}^t$ and $\mathcal{G}^t$. If there is no $\mathcal{D}^t$, all the generation are examined only by local discriminator in each client. Then the generation from different clients will not be correlated because their correlations are never checked. If there is no $\mathcal{G}^t$, each of the client will sample the random noise vector separately instead of using a common one, then there is no guarantee that the synthetic data can learn the column correlations across different clients.
One may also question the necessity of $\mathcal{D}^t$ and $\mathcal{G}^t$. Without $\mathcal{D}^t$, all generations would be examined only by discriminators in each client. As a result, the generations from different clients would not be correlated because their correlations are never checked. Similarly, without $\mathcal{G}^t$, each client would sample a separate random noise vector for $\mathcal{G}^b_1$ and $\mathcal{G}^b_2$, which would prevent the synthetic data from learning column correlations across clients since there is no correlation between two separate random noise vectors.

\begin{algorithm}[h]
    \SetAlgoLined
    \KwIn{Number of client $N$; Server side generator $\mathcal{G}^t$ and discriminator $\mathcal{D}^t$; Conditional vector filter $\mathcal{D}^s$; Client side generator $\mathcal{G}^b_i$, discriminator $\mathcal{D}^b_i$ and data $T_i$ for client $i$, $i\in[1,N]$; Training round $R$. Discriminator training epochs per round $e$.}
    \KwOut{Trained Generator $\{\mathcal{G}^t$, $\mathcal{G}^b_i\}$, Discriminator $\{\mathcal{D}^t$, $\mathcal{D}^s$, $\mathcal{D}^b_i$\}}
    % Training of Discriminator\;
    \nonl\textbf{Client $i$ ($i\in[1,N]$) operates:}\\
    \texttt{Encode}($T_i$)\;
    \While{ current round $< R$}{
        \nonl\textbf{Train Discriminator, freeze $\{\mathcal{G}^t$, $\mathcal{G}^b_i\}$:}\\
        \While{current local epoch $< e$}
        {
          \nonl \textbf{Server operates:}\\
           $CV_p$, $idx_p$ <- \texttt{CVGeneration}(N clients) where the condition column is constructed by Client $p$\\
            \{$\mathcal{G}^t_{i}(Z, CV_{p}), i\in[1,N]$\} <- \texttt{Split}$(\mathcal{G}^t(Z, CV_p))$\\ 
           \nonl\textbf{Client $i$ ($i\in[1,N]$) operates:}\\
           Output $\mathcal{D}^b_i(\mathcal{G}^b_i(\mathcal{G}^t_{i}(Z, CV_p)))$ as $\mathcal{D}^{b,i}_{out}$ to server\;
          \nonl\textbf{Server operates:}\\
           $\mathcal{D}^t_{inS}$ <- \texttt{Concat}($\mathcal{D}^{b,i}_{out}$, $\mathcal{D}^s(CV_p)$), $i\in[1,N]$)\;
           Output $\mathcal{D}^t(\mathcal{D}^t_{inS})$\;
           \nonl\textbf{Client $i$ ($i\in[1,N]$) operates:}\\
            \eIf{$i == p$}{
                Output $\mathcal{D}^b_p(T_p[idx_p])$ to server\;
            }{
                Output $\mathcal{D}^b_i(T_i)$ to server\;
            }
           \nonl\textbf{Server operates:}\\  
           $\mathcal{D}^t_{inR}$ <- \texttt{Concat}($\mathcal{D}^b_i(T_i)[idx_p]$, $\mathcal{D}^b_p(T_p[idx_p])$,  $\mathcal{D}^s(CV_p)$), $i\in[1,N]$ and $i$ $!=$ $p$)\;
           Output $\mathcal{D}^t(\mathcal{D}^t_{inR})$\;
           Calculate \texttt{Loss$^\mathcal{D}$}($\mathcal{D}^t(\mathcal{D}^t_{inS})$, $\mathcal{D}^t(\mathcal{D}^t_{inR})$) and update $\{\mathcal{D}^t$, $\mathcal{D}^s$, $\mathcal{D}^b_i$\}.
        }  
        \nonl\textbf{Train Generator, freeze $\{\mathcal{D}^t$, $\mathcal{D}^s$, $\mathcal{D}^b_i$\}}\\
        \nonl\textbf{Server operates:}\\
        $CV_p$, $idx_p$ <- \texttt{CVGeneration}(N clients) where the condition column is constructed by Client $p$\\
        \{$\mathcal{G}^t_{i}(Z, CV_{p}), i\in[1,N]$\} <- \texttt{Split}$(\mathcal{G}^t(Z, CV_p))$\\ 
        \nonl\textbf{Client $i$ ($i\in[1,N]$) operates:}\\
        Output $\mathcal{D}^b_i(\mathcal{G}^b_i(\mathcal{G}^t_{i}(Z, CV_p)))$ as $\mathcal{D}^{b,i}_{out}$ to server\;
        \nonl\textbf{Server operates:}\\
        $\mathcal{D}^t_{inS}$ <- \texttt{Concat}($\mathcal{D}^{b,i}_{out}$, $\mathcal{D}^s(CV_p)$), $i\in[1,N]$)\;
        %\wh{symbol for synthetic data?}
        Calculate \texttt{Loss$^\mathcal{G}$}($\mathcal{D}^t(\mathcal{D}^t_{inS})$) and update $\{\mathcal{G}^t$, $\mathcal{G}^b_i\}$.\;
        \nonl\textbf{Client $i$ ($i\in[1,N]$) operates:}\\
        \texttt{Shuffle($T_i$)}, $i\in[1,N]$\;
    } 
\caption{\algo Training Process}
\label{alg1}
\end{algorithm}

\subsubsection{{Threat Model}} \label{ssec:threat_model}
In line with many previous works in the field of VFL \cite{wu2020privacy, fang2021large, fu2021vf2boost, fu2022blindfl}, we focus on the semi-honest model in the design of our proposed \algo architecture. {Specifically, we assume that the clients and server are \textit{honest but curious}, and their behaviors obey the following rules:}

\begin{enumerate}[leftmargin=*]
    \item {All clients and the server comply with the \algo protocol and execute the training process honestly.}
    \item {While any client or server may attempt to extract additional information from the exchanged messages, none of them intentionally modify the values to disrupt the \algo training process.}
    \item {Clients do not engage in any collusion to exchange information with other clients or the server except for necessary messages.}
\end{enumerate}

{These assumptions form our threat model for the \algo architecture, and our design aims to address potential threats that may arise under these assumptions.}

% meaning that they adhere to the GTV protocol, but may seek to acquire additional information through the computation process. We do not consider collusion between server and clients. Despite the fact that tabular datasets are retained locally within each client, the values derived during the GTV training and synthesizing process may also reveal sensitive information and result in data leakage.} 

\subsubsection{\algo Training Procedure}
The training process of \algo is similar to that of a centralized tabular GAN, despite the fact that \algo distributes the default tabular GAN into multiple separate components. As the example shown in Fig.~\ref{fig:gtv_workflow}, the workflow is numerated, parallel actions are noted with the same numbers.

During the discriminator training phase, the process starts with the generator. After initializing the random noise and conditional vectors, they are concatenated and fed to $\mathcal{G}^t$ and $\mathcal{D}^s$ (i.e., \circled{1}). The data is then passed through \circled{2} to \circled{5}. After  \circled{5}, the server concatenates the intermediate logits and uses the concatenation as input for $\mathcal{D}^t$ to produce predictions for the synthetic data. At the same time, clients select their real training data following  \circled{6} then \circled{5}, and the server concatenates the intermediate logits and produces predictions for the real data. These predictions for real and synthetic data are used to calculate the gradients for updating $\{\mathcal{D}^t$, $\mathcal{D}^s$, $\mathcal{D}^b_1$, $\mathcal{D}^b_2$\}.

During the generator training phase, the process also begins with the generator. The concatenation of the random noise and conditional vectors is passed through \circled{1} to  \circled{5}, and $\mathcal{D}^t$ produces predictions. These predictions are used to update $\{\mathcal{G}^t$, $\mathcal{G}^b_1$, $\mathcal{G}^b_2\}$. Once all network components are updated, one training round is complete. At the end of each training round, all clients shuffle their local training data using the same random seed to ensure data consistency between them.

\subsubsection{Training Details}
\label{sssec:training_details}
%\lc{The description need to be matched with the line of algorithm. Add the line numbers behind the description consistently.}
Algo.~\ref{alg1} elaborates the training flow detail. The whole process involves of several functions, namely \texttt{Encode}, \texttt{CVGeneration}, \texttt{Split}, \texttt{Concat}, \texttt{Loss$^\mathcal{D}$}, \texttt{Loss$^\mathcal{G}$} and \texttt{Shuffle}. In step \textbf{1}, each client needs to encode their local data. One-hot encoding for categorical column, mode-specific normalization~\cite{ctgan} for categorical column and mixted-type encoder~\cite{ctabgan} for column contains both categorical and continuous values. In step \textbf{3}, the training of discriminator needs a local training epochs $e$ because our tabular GAN algorithm is based on Wasserstain GAN plus gradient penalty~\cite{wgan_gp} loss function. This training method needs more discriminator update (by default $e$ = 5) than generator; In step \textbf{4}, server uses \texttt{CVGeneration} to notify all the clients to generate their local conditional vector (CV) based on the CV construction method proposed by \ctgan~\cite{ctgan}, then it randomly choose one of the clients' CVs based on the probability vector $P_r$ calculated by the ratio of number of features in the client to the total number of features across all the clients. The selected client $p$ does not only upload the $CV_p$ to the server, but also the indices of selected training data $idx_p$ that meets the conditions represented in $CV_p$.
%and upload to the server. 
%the results for the current round. 
Due to privacy concern, the $idx_p$ is only kept between the selected client and server, and the potential risk is discussed in Sec.~\ref{ssec:further_discussion}.
%, otherwise other clients can reconstruct the data columns of the client $p$ by the $CV_p$;  
In step \textbf{5}, \texttt{Split} function vertically splits output logits proportionally based on the ratio vector $P_r$.
% that server calculated before; 
In step \textbf{7}, \texttt{Concat} function horizontally concatenates all the intermediate logits into one input $\mathcal{D}^t_{inS}$. During step \textbf{9} to \textbf{13}, for the client $p$, it just needs to select out the $idx_p$ training data and passes through the  $\mathcal{D}^b_p(T_p[idx_p])$. For the rest of the clients, they need to pass all their local data through the local discriminator; Then in step \textbf{14}, except the output from client $p$, the server needs to first use $idx_p$ to select their outputs, then concatenate all these intermediate logits into one input $\mathcal{D}^t_{inR}$. In step \textbf{16}, With $\mathcal{D}^t_{inS}$ and $\mathcal{D}^t_{inR}$, we can use \texttt{Loss$^\mathcal{D}$} to calculate gradients and update all discriminator parts. The training steps for the generator from \textbf{18} to \textbf{21} are the same as the steps from \textbf{4} to \textbf{7}. The difference is that to update the generator, function \texttt{Loss$^\mathcal{G}$} (i.e., the step \textbf{22}) only needs $\mathcal{D}^t_{inS}$. 

% \zz{add a figure for shuffling}

\begin{figure}[t]
    \centering
    \includegraphics[width=0.95\linewidth]{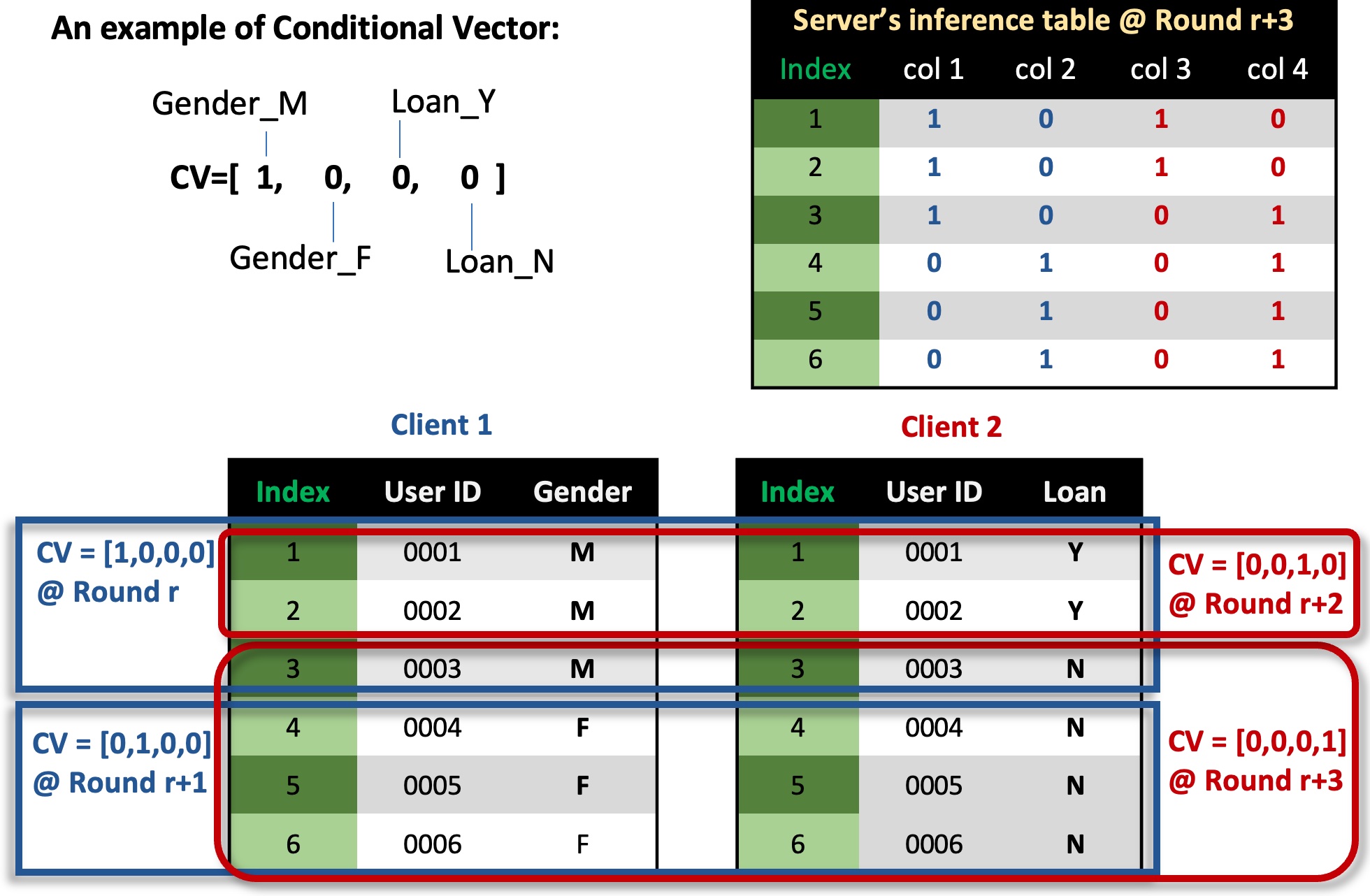}
    %\vspace{-1em}
    \vspace{-0.5em}
    \caption{\algo training without shuffling.}
    \label{fig:noshuffle}
   \vspace{-1.3em}
\end{figure}

\subsubsection{Training-with-shuffling} 
\label{sssec:training-with-shuffling}
At the end of each training round (i.e., step \textbf{23} in Algo.~\ref{alg1}), we need to call  \texttt{Shuffle} on all the clients in order to shuffle their local training data. Moreover, all the clients share the same random seed inside their \texttt{Shuffle} function, which means after \texttt{Shuffle}, each data row across all the clients is still aligned to the same ID, only the data index changes. We assume that \texttt{Shuffle} function and random seed are securely negotiated among all the clients before training and isolated from server. 
%We also assume that the \texttt{Shuffle} function and random seed in client side are isolated from server. 
% Otherwise with the training data index and chosen category in conditional vector, server can reconstruct clients' data.

Fig.~\ref{fig:noshuffle} shows an example of \algo training WITHOUT \texttt{Shuffle} in step \textbf{23} in Algo.~\ref{alg1}. Assuming the system contains two clients and each client contains only one feature, i.e., Gender and Loan, respectively. Both features are categorical columns and each of them has only two classes. According to the CV construction method of \ctgan, the CV of our example  contains four bits, each of which represents one class of the categorical columns, and we can only indicate one class per CV. For each training round, we sample new CVs, and the corresponding classes of selected local training data need to match the new CVs. For example, at round $r$, if our sampled CVs contain three [1,0,0,0] vectors, then the corresponding data indices (1,2,3) also return to the server (i.e., step \textbf{4} in Algo.~\ref{alg1}), then round $r+1$ with three [0,1,0,0] in CVs and indices (4,5,6), round $r+2$ with two [0,0,1,0] in CVs and indices (1,2), and finally round $r+3$ with four [0,0,0,1] in CVs and indices (3,4,5,6). 
{The server can use the indices and CVs obtained during the training process to reconstruct the entire dataset after a certain number of rounds, as illustrated in Figure~\ref{fig:noshuffle}. Despite not having knowledge of the number of categorical columns or their names in the system, the server can infer this information by analyzing the one-hot encoding of each row. In Figure~\ref{fig:noshuffle}, after round $r+3$, the server can deduce that there are two categorical columns in the system, each containing two categories. Additionally, the server can infer the ratio of categories within each column, such as a 1:1 ratio for the first column and a 1:2 ratio for the second column. If the synthetic data is released by clients, the server can determine the corresponding column names by analyzing the ratio of categories in all categorical columns.}

\begin{figure}[t]
    \centering
    \includegraphics[width=0.88\linewidth]{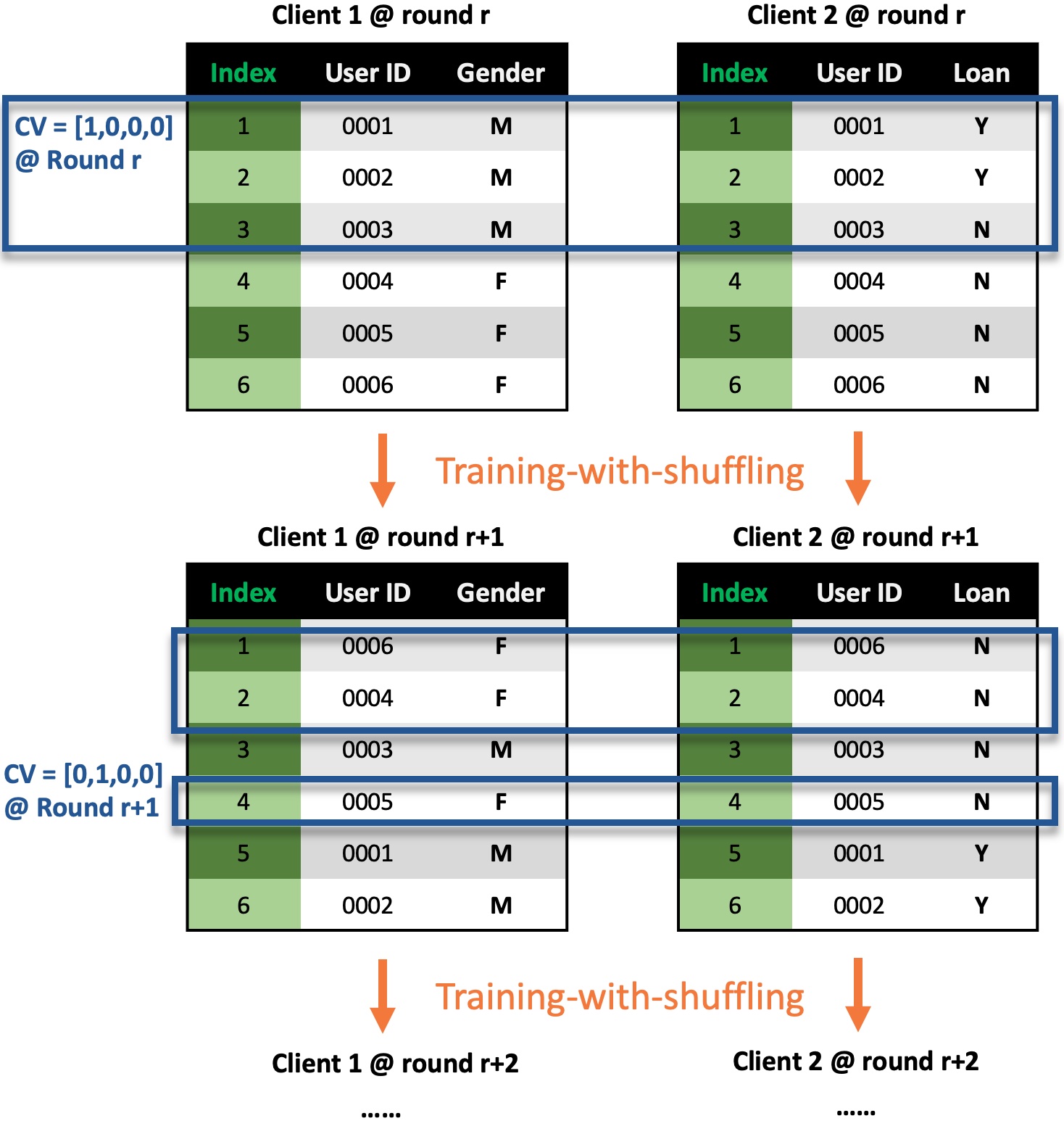}
    \vspace{-0.5em}
    \caption{\algo training-with-shuffling.}
    \label{fig:shuffle}
   \vspace{-1.3em}
\end{figure}

Fig.~\ref{fig:shuffle} shows an example of the \algo with \textit{training-with-shuffling}. At round $r$, we still sample CVs that contain three [1,0,0,0] vectors and get the corresponding data indices (1,2,3). But this time, we launch the shuffling function in each client. At round $r+1$, we can see that the index column remains the same as last round but the data content part have shuffled, and more important the client 1 and client 2 have the same order in User ID column. 
%This is because we use the same random seed for \texttt{shuffle} function in all the clients.  
If round $r+1$ sampled CVs that contains three [0,1,0,0], the server gets the indices [1,2,4]. But these indices in the new round corresponds to different data content. Therefore, the server cannot reconstruct clients' data.

\subsubsection{Further Discussion}
\label{ssec:further_discussion}
% Accommodating CV into \algo training is not easy. Since CV is part of the input of generator, we cannot hide it from server. But we have the control of passing or not the selected data indices to server. \algo chooses to share this information with server, therefore, at step \textbf{12}, we let the clients who did not contribute CV to pass the whole local data through $\mathcal{D}^b_i$ and then at step \textbf{14} to select the corresponding indices data from the intermediate logits in the server. This design increases the local calculation and the communication between server and clients, but it makes sure no privacy leakage. 
Incorporating CV into the training of the \algo algorithm is not a straightforward task. Since CV is a part of the generator's input, it cannot be hidden from the server. However, we have control over whether to pass selected data indices to the server. \algo chooses to share this information with the server, so at step \textbf{12}, we allow clients who do not contribute CV to pass their entire local data through $\mathcal{D}^b_i$, and then at step \textbf{14}, the server selects the corresponding indices data from the intermediate logits. This design increases local computation and communication between the server and clients, but it ensures that there is no privacy leakage. {In order to decrease the overall training time, certain calculations can be performed in parallel. After step \textbf{4}, all the clients are aware of whether their conditional vectors have been selected. Therefore, all the clients are able to execute steps \textbf{10} or \textbf{12} right after step \textbf{4}.}

% One may argue that what if client did to not share the selected data indices to server, instead shares this information only among all the clients in a peer-to-peer (P2P) manner. In that case, server does not have the indices information, then we do not need to shuffle local data and all the clients only need to pass the selected data rows to pass their local $\mathcal{D}^b_i$. First of all, maintaining P2P connections among all the clients can be expensive and insecure, the number of P2P connections increases exponentially with the increase of number of clients. Furthermore, this solution also leaks clients' privacy in an indirect way. In order to over-sample the minority class in imbalanced column, the CV of \ctgan is sampled by the log-frequency of the class appearances in the column. Since it is over-sampled, there will have the scenario that several data rows who contain minority class are selected multiple times in one training epoch. These indices are also often selected together. Then for the client who did not contribute CV in current training epoch can rationally guess that the data of these indices has some common features in a certain column of the CV contributor. This is also a privacy leakage. 

An alternative approach to design \algo is to make clients share the selected data indices in a peer-to-peer (P2P) manner rather than with the server. In this case, the server would not have access to the indices information, and clients would only need to pass the selected data rows through local $\mathcal{D}^b_i$. However, this solution has several limitations. Firstly, maintaining P2P connections among all the clients can be costly and insecure, as the number of connections increases exponentially with the number of clients. Additionally, this approach also indirectly leaks clients' privacy.
To balance the class imbalances in the columns of real data, the CV of \ctgan is sampled based on the log-frequency of class appearances in the column. As a result, there may be scenarios where several data rows containing the minority class are selected multiple times in one training epoch, often together. In this case, a client who does  not contribute CV in the current training round could reasonably infer that the data in these indices have common features in a certain column of the CV contributor, which causes a privacy leakage. Adding shuffling to this design cannot solve above inference. Because in VFL, we can only shuffle the index of data, the mapping of other features among all clients remain the same (see Fig.~\ref{fig:shuffle}), all the clients can observe the repeated selected group of data if it exits.

\subsubsection{{Synthetic Data Publication}} 
\label{ssec:data_publication}
% \lc{move this to the last section of workflow}
{Once the training is complete, the secure publication of synthetic data becomes a crucial step in the \algo framework. In practical scenarios, the server receives a request for a specific amount of synthetic tabular data and responds by sending an equivalent amount of random noise to the generator $\mathcal{G}$. The clients independently use their own generators $\mathcal{G}^b_i$ to generate their respective synthetic data portions, which are then concatenated to form the final synthetic dataset.}

{To ensure privacy preservation, clients must shuffle their synthetic data before publication. Without shuffling, the server would have partial black-box access \cite{chen2020gan} to $\mathcal{G}$, which assumes knowledge of the inputs (random noise) and outputs (published synthetic data), including the input-output pairs. This partial black-box access poses significant privacy risks in \algo, enabling independent attacks by the server without client collusion. Hence, shuffling before publishing is essential to eliminate partial black-box access, preventing the server from possessing knowledge of the input-output pairs of $\mathcal{G}$.} 
{It is important to note that in \algo, nothing beyond the final synthetic data is published. The partial neural networks employed by the clients and server are considered private assets and are not deployed as part of the API.}

% This is because in step \circled{3}, the server has an access to the inputs of $\mathcal{G}^b_i$ and $\mathcal{G}^b_i$, and the synthetic data is the output of these generators. Without shuffling, the server can use this information to reconstruct the bottom generators. With shuffling, there is no input and output mapping for bottom generators that server can observe.

% \zz{Should insert an algorithm or a training process table?}

\subsection{Privacy Analysis in \algo Training Flow} 
%\lc{The privacy formulation is not an accurate name. Suggestion: privacy enhancing GTV training flow}

\label{ssec:privacy_form}
% \subsubsection{Threat Model}
%In line with many previous works in the field of VFL \cite{wu2020privacy, fang2021large, fu2021vf2boost, fu2022blindfl}, we focus on the semi-honest model in the design of our proposed GTV architecture. Specifically, we assume that the clients and server are honest but curious, meaning that they adhere to the GTV protocol, but may seek to acquire additional information through the computation process. Despite the fact that tabular datasets are retained locally within each client, the values derived during the GTV learning process may also reveal sensitive information and result in data leakage. 
According to threat model, here we analyze the privacy leakage risk in \algo training, and explain the motivations behind our design. We highlight the sensitive information exchanged and used in \algo, including feedforward activations, and backpropagation process on $\mathcal{G}$ and $\mathcal{D}$, model weights, and model gradients. Here both $\mathcal{G}$ and $\mathcal{D}$ represent the combination of top and bottom generator and discriminator in \algo.

\textbf{Feedforward in Generator.} The input data of this process is a combination of random noise $Z$ and conditional vector $CV$, which is passed through the neural network layers of $\mathcal{G}$ until it reaches the output layer on each client. Finally, the corresponding piece of synthetic tabular data is generated on each client as the output. Obviously, none of the participants is able to reconstruct real records during this process since the real tabular datasets $T_{i}$ are not used throughout this process. Even if all clients except one collude with the server to infer information about the remaining client, the inferred information is only about synthetic data, which is actually intended to be published in our design. 

\textbf{Backpropagation in Generator.} In this process the generator $\mathcal{G}$ takes the loss \texttt{Loss$^\mathcal{G}$}($\mathcal{D}^t(\mathcal{D}^t_{inS})$) as input and updates its model parameters using the Stochastic Gradient Descent (SGD) method. Leakage of labels is a significant concern in the backpropagation process of traditional VFL classification models\cite{fu2022label,fu2022blindfl}. However, this risk does not apply for $\mathcal{G}$ because (i) the backpropagation of $\mathcal{G}$ does not involve the use of real dataset $T_{i}$, as outlined in Section \ref{sssec:training_details}; (ii) $\mathcal{G}$ is not a classifier, and thus, the concept of 'label' is not applicable. It is possible for the client to infer the $CV$ by examining changes in gradients\cite{fu2022label}. Nonetheless, as elaborated in Section \ref{sssec:training-with-shuffling}, only the selected client knows which rows are selected based on its CV. Neither server nor other clients can reconstruct selected client's data only by CV. 
%Those clients who are not selected are requested to upload the results of all rows thus they cannot infer other clients' data even given the value of CV. 
Therefore, the leakage of CV is not considered as a significant privacy risk.

% since the clients, who do not contribute CV, do not have the information of chosen training data indices, they cannot deduce other clients' data only according to CV. This is not considered a significant privacy risk.
% as elaborated in Section \ref{sssec:training-with-shuffling} this is not considered a significant privacy risk.
%\zz{if client knows CV, the client can reconstruct other client's data. I think it is a problem.}.

\textbf{Feedforward in Discriminator.} 
% In contrast, t
The architecture of discriminator $\mathcal{D}$ aligns more closely to a traditional VFL classifier. The main difference lies in the labels of the training data. In the training process of a VFL classifier, the label column is held by one of the clients, who also manages the server, as illustrated in Fig.~\ref{fig:vfl_classifier}. 
In \algo, the labels for training $\mathcal{D}$ are only \textit{real} and \textit{fake}, which are not part of the original tabular dataset. Additionally, it is each client who labels \textit{real} and \textit{fake} for the data instances. Therefore, the threat models mentioned in \cite{fu2022label, li2022label, fu2022blindfl} do not apply for $\mathcal{D}$, since they all assume that a malicious client, or multiple clients, attempt to infer the label column which they do not possess. 

The intermediate logits sent from clients to the server may contain sensitive information about the dataset. An example of this is mentioned in \cite{fu2022blindfl}, which points out that the server can determine if two records have the same categorical values by comparing their intermediate logits during the feedforward process. 
%Privacy risks like this are mitigated in GTV 
\algo mitigates the risk like that
because the server is independent from all the clients and it does not hold any information about the dataset.

% \zz{is that phrase repeating?}
% It should also be noted that the server in GTV is not the holder of the \textit{fake} and \textit{real} labels and is not affiliated with any of the clients. Thus the server should be prohibited from obtaining any information about the labels. This can be guaranteed by the \textit{training-with-shuffling} approach mentioned in Section \ref{sssec:training-with-shuffling}. An example of this is mentioned in \cite{fu2022blindfl}, which points out that the server can determine if two records have the same categorical values by comparing their intermediate logits during the feedforward process. Privacy risks like this are mitigated in GTV benefiting from the \textit{training-with-shuffling} process. This is because the server is unaware of the true order of training data thus unable to gain knowledge about the profile of the training dataset. 

% Note that in the extreme case that a malicious client colludes with the server, the risk may exist but 

\textbf{Backpropagation in Discriminator.} As mentioned above, the clients already know the \textit{real} and \textit{fake} labels of the training data prior to the training of $\mathcal{D}$, thus the label leakage problem \cite{fu2022blindfl} of the backpropagation process is not an issue in \algo. One potential risk is that malicious clients may collude with the server to infer sensitive information about the training data of the victim client. However, it is important to note that this information, such as distributions and imbalanced features, is an inherent aspect of the final product of \algo, which is a joint synthetic tabular dataset. Therefore, it can be argued that this risk is justified in the application scenario of \algo, as clients who participate in the program are aware of such risks.
%and choose to proceed with the training process.

\textbf{Model weights and model gradients.} The model weights are considered intensive information in VFL studies \cite{wu2020privacy,fu2022blindfl}, since the values indicate the relative importance of each feature for the learning tasks. In \algo, this may cause feature importance leakage once the malicious clients share the trained model weights of $\mathcal{D}$ with the server. As mentioned above, the feature importance here is only in reference to the impact of the features on distinguishing real and synthetic data. Additionally, while traditional VFL tasks take the trained model as the final product and it is normal to share the model weights among clients, our \algo sees the joint synthetic dataset as the ultimate outcome. As a result, model weights are supposed to be kept safe and never exposed in \algo.
%by each participant in GTV.

Recent studies \cite{zhu2019deep,geiping2020inverting,jin2021cafe} have shown that model gradients can lead to privacy leakage. In \cite{jin2021cafe}, the authors present a new algorithm CAFE for recovering large amounts of private data from shared aggregated gradients in the VFL settings. CAFE has a high recovery quality and is backed by theoretical guarantees. However, the assumptions made by the algorithm are quite restrictive and may not be practical for our \algo scenarios. One of the key prerequisites in \cite{jin2021cafe} is that the server knows the data index, which is not feasible in our settings. Just as with the privacy of model weights, the risk of data leakage from gradients can only occur when malicious clients are working in collusion with the server.

In summary, when malicious participants act alone, the \algo system is robust to privacy leakage issues present in traditional VFL. However, the system becomes vulnerable when multiple parties collude. This can lead to even more severe security problems, as other advanced attacks may also pose a threat to \algo. These issues are discussed in greater detail in the following section.

\subsection{Security Analysis} \label{ssec:security_analysis}
%\lc{We are not security experts. Statements below are rather strong. Can you tone down them? I change the name to Risk analysis}
Here, we discuss the adversarial attacks in the context of  \algo.

\textbf{Feature inference attack.} For the feature inference attacks in VFL, Luo et al. \cite{luo2021feature} first propose attacks on the logistic regression (LR) and decision tree (DT) models and achieve high attack accuracy. However, these attacks are not applicable to complex models such as neural networks. In our \algo architecture, both $\mathcal{G}$ and $\mathcal{D}$ are neural network models characterized by the presence of multiple hidden layers that facilitate non-linear transformations. This structure poses a natural barrier to equality-solving attacks based on solving equations and the maximum a posteriori (MAP) method \cite{fredrikson2015model}.
Luo et al. \cite{luo2021feature} then design a generative regression network (GRN) to learn the correlations between the malicious client's and victim client's features. Even for complex neural network models, GRN manages to infer the distribution of feature values with high accuracy. However, this method relies on the assumption that the trained VFL model is given to the attacker. As discussed in the previous section, although publishing the trained models is the common case in VFL scenarios, it is prohibited in \algo. In other words, conducting \cite{luo2021feature}'s feature inference attack on \algo requires the attacker to steal model parameters of $\mathcal{G}$ and $\mathcal{D}$ from all participants, which is unrealistic.

\textbf{Label inference attack.} Label leakage problem has been widely studied for VFL \cite{vepakomma2019reducing, li2022label, fu2022label}. Although the results show that the malicious client succeeds to infer some label information, these studies are case-specific and have some limitations. \cite{li2022label} works only for the case of binary classification model training between two VFL clients. \cite{fu2022label} is based on the assumption that the adversary has a small amount of auxiliary labelled data to fine-tune the attack model. 
Moreover, as discussed in Section \ref{ssec:privacy_form}, the \textit{real} and \textit{fake} label in \algo is not part of the tabular original dataset and is managed by clients. To the best of our knowledge, such label inferences attacks on \algo do not make much sense as in traditional VFL cases. 

% However, it is important to consider the potential for attacks on the sensitive information contained in the \textit{real} and \textit{fake} labels, which have been studied in the context of attacks on generative adversarial networks (GANs). This perspective on the issue is discussed further below.

% \textbf{Backdoor attack.} This is not related to privacy, so we do not necessarily need to talk about it in this paper.

\textbf{Membership Inference Attack.} {In addition to the aforementioned privacy threats on VFL, a distinct form of attack, known as the Membership Inference Attack (MIA), specifically targets generative models such as GANs \cite{hilprecht2019monte, chen2020gan, hu2021tablegan}. The MIA is executed in the following manner: the adversary is granted access to the machine learning model, which could be black-box or white-box access depending on the specific scenario. The adversary also has access to a set of data records. By conducting MIA, the adversary attempts to infer whether a data record has been used to train the model. This could potentially expose sensitive information.}
%\wh{need to explain MIA here, and mention our experiments}
%, which is typical in centralized GANs. 

{In the structure of \algo, the generator $\mathcal{G}$ is partitioned into top and bottom models, each owned by separate participants. Consequently, it appears implausible to posit that an attacker can gain white-box access to the entire $\mathcal{G}$, as this presumes a full collusion between all clients and the server. As discussed in Section \ref{ssec:data_publication}, the shuffling mechanism prior to publishing mitigates partial black-box access to $\mathcal{G}$. Nonetheless, under our threat model, the possibility remains for an adversary to conduct MIA with full black-box access, which means the attacker is only permitted to access the generated samples set from the well-trained black-box $\mathcal{G}$. In Section~\ref{sss:dp_gtv}, we execute \cite{chen2020gan}'s full black-box MIA on \algo, thereby assessing the effectiveness of the privacy protection measure - Differential Privacy (DP), in counteracting this attack.}

\section{Empirical Evaluation}
\label{sec:experiment}
\subsection{Experiment Setup}
\label{sec:experiment_setup}
\textbf{Datasets}
Our algorithm is tested on five commonly used machine learning datasets: Adult, Covertype, Intrusion, Credit, and Loan. Adult, Covertype, and Intrusion are obtained from the UCI machine learning repository\footnote{\url{http://archive.ics.uci.edu/ml/datasets}}, while Credit and Loan are obtained from Kaggle\footnote{\url{https://www.kaggle.com}}. All five datasets contain a target variable. Although it is not required for a tabular dataset to have a target column for data synthesis, we have chosen to use datasets with a target column for the purpose of classification evaluation. This allows us to test the utility of the synthetic data for machine learning tasks compared to real data. Due to computing resource limitations, we randomly sample 50K rows of data in a stratified manner with respect to the target variable for the Covertype, Credit, and Intrusion datasets. The Adult and Loan datasets are not sampled. For all the datasets, we train the algorithms 300 epochs for all the experiments. Each experiment is repeated three times and the average result is reported. % It is worth noting that target columns of all five datasets are imbalanced.

\textbf{Baseline}
To the best of our knowledge, \algo is the first architecture to incorporate SOTA tabular GAN into vertical federated learning. Our baseline for comparison is therefore a \textbf{centralized tabular GAN}. To have a better centralized baseline, we have adopted various techniques from \ctgan~\cite{ctgan} and CTAB-GAN~\cite{ctabgan}. These include the use of one-hot encoding for categorical columns, mode-specific normalization for continuous columns, mixed-type encoder for columns containing both categorical and continuous values. The construction method of the conditional vector is also adopted from \ctgan. All of these techniques are incorporated into our centralized tabular GAN. 
% The generator and discriminator structures are adopted from \ctgan. 
The generator and discriminator structures are the same as in \ctgan. The generator is a Resnet-style~\cite{he2016deep} neural network comprising two residual blocks followed by a fully-connected (FC) layer. Each residual network (RN) block includes three successive layers: FC, batchnorm, and relu. The discriminator has two fully-connected network (FN) blocks followed by another FC layer. Each FN block consists of three successive layers: FC, leakyrelu, and dropout. The output dimension for all the RN block and FN block is fixed to 256.

\textbf{Testbed}
The experiments are conducted on two machines running Ubuntu 20.04. Both machines are equipped with 32 GB of memory, an NVIDIA GeForce RTX 2080 Ti GPU, and a 10-core Intel i9 CPU. One machine serves as the server, while the other hosts all the clients range, with the number of clients ranging from 2 to 5. 
%\lc{How many clients}

% \begin{figure*}[t]
% 	\begin{center}
% 		\subfloat[Generator]{
% 			\includegraphics[width=0.48\textwidth]{figures/generation.jpg}
% 			\label{fig:generator}
% 		}
% 		\subfloat[Discriminator]{
% 			\includegraphics[width=0.48\textwidth]{figures/discriminator.jpg}
% 			\label{fig:discriminator}
% 		}
%             \vspace{-1em}
% 		\caption{Neural Network Partition} 
% 		\label{fig:partition_structure}
%  	\end{center}
%  \vspace{-1em}
% \end{figure*}

\begin{figure*}[t]
	\begin{center}
		\subfloat[Generator]{
			\includegraphics[width=0.47\textwidth]{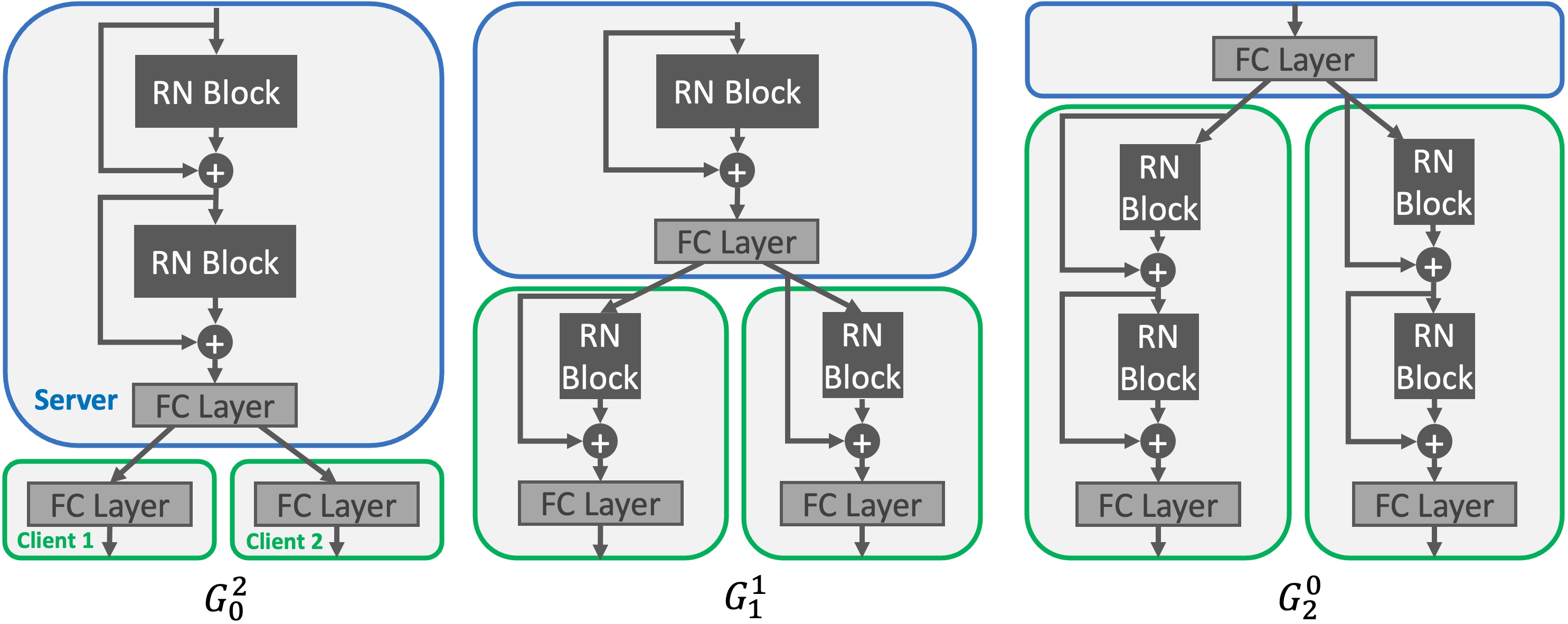}
			\label{fig:generator}
		}
		\subfloat[Discriminator]{
			\includegraphics[width=0.47\textwidth]{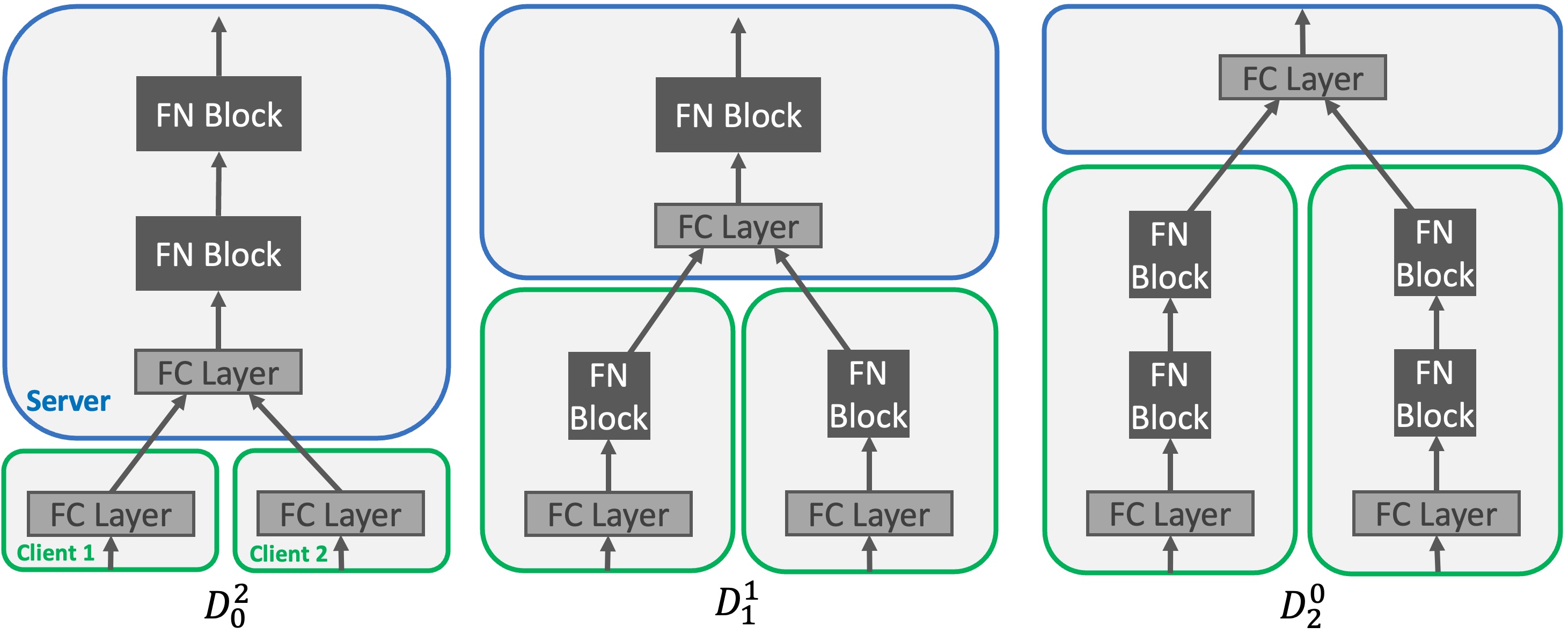}
			\label{fig:discriminator}
		}
            \vspace{-1em}
		\caption{Neural Network Partition. $G^{n_1}_{n_2}$ denotes the generator with $n_1$ RN block(s) in server and $n_2$ RN block(s) in each client. $D^{n_3}_{n_4}$ denotes the discriminator with $n_3$ FN block(s) in server and $n_4$ FN block(s) in each client.}
  %\lc{In the figure: they are good place to provide the semantic meaning of this G and D.}} 
		\label{fig:partition_structure}
 	\end{center}
 \vspace{-1em}
\end{figure*}

\subsection{Evaluation Metrics}
We evaluate the synthetic data in two dimensions: (1) \textbf{Machine Learning (ML) Utility} and (2) \textbf{Statistical Similarity}. The two dimensions cover the row-wise correlation and column-wise distribution of the synthetic data. Given the local training data $RD_A$ and $RD_B$ for clients A and B, respectively, the synthetic data produced by the clients is denoted as $SD_A$ and $SD_B$. Unless otherwise stated, in the following context, synthetic data refers to the concatenation of $SD_A$ and $SD_B$, and real data refers to the concatenation of $RD_A$ and $RD_B$. % It is important to note that the metrics presented in the following sections are reported as the difference between real and synthetic data. Therefore, a lower value indicates better quality of the synthetic data. 
%\lc{We need a higher number of clients. Two clients is too low}

\subsubsection{Machine Learning Utility}
To evaluate the effectiveness of synthetic data for machine learning tasks, we have designed the following evaluation pipeline. We first split the original dataset into training and test sets. The training set (i.e., $RD_A$ and $RD_B$) is used as input to the \algo models to generate synthetic data of the same size. The synthetic and real training data are then used to train five widely used machine learning algorithms (decision tree classifier, linear support vector machine, random forest classifier, multinomial logistic regression, and multi-layer perceptron). The trained models are then evaluated on the real test set. Performance is measured using accuracy, F1-score, and the area under the curve (AUC) of the receiver operating characteristic curve, difference between models trained by real and synthetic data is reported. The goal of this design is to determine whether synthetic data can be used as a substitute for real data to train machine learning models. 

\subsubsection{Statistical Similarity}
Three metrics are used to quantify the statistical similarity distance between real and synthetic data. 

\textbf{Average Jensen-Shannon divergence (JSD)}. The JSD is a measure of the difference between the probability mass distributions of individual categorical columns in the real and synthetic datasets. It is bounded between 0 and 1 and is symmetric, making it easy to interpret. We average the JSDs from all the categorical columns to obtain a compact, comprehensible score.

\textbf{Average Wasserstein distance (WD)}. 
%To measure the similarity between the distributions of continuous/mixed columns in synthetic and real datasets, we use the Wasserstein distance. 
We find that the JSD metric is numerically unstable for evaluating the quality of continuous columns, especially when there is no overlap between the synthetic and original datasets, so we chose to use the more stable Wasserstein distance to measure the similarity between the distributions of continuous/mixed columns in synthetic and real datasets. We average all column WD scores to obtain the final score.
 
\textbf{Difference in pair-wise correlation (Diff. Corr.)}.
To evaluate the preservation of feature interactions in synthetic datasets, we compute the pair-wise correlation matrix separately for real and synthetic datasets. 
% The \textit{Pearson correlation coefficient} is used between two continuous columns, ranging from -1 to 1. The \textit{Theil uncertainty coefficient} is used between two categorical columns, ranging from 0 to 1. The \textit{correlation ratio} is used between categorical and continuous columns, ranging from 0 to 1. 
The dython\footnote{\url{http://shakedzy.xyz/dython/modules/nominal/\#compute\_associations}} library is used to compute correlation matrix for each table. 
%\lc{The following description is difficult to follow. You should write first why you need those metrics and then explain them later } 
The $\mathit{l}^2$-norm of difference between the real and synthetic correlation matrices is then calculated and abbreviated as \textbf{Diff. Corr.}.
% To evaluate how well feature interactions are preserved in the synthetic datasets, we first separately compute the pair-wise correlation matrix for real and synthetic datasets.
% \textit{Pearson correlation coefficient} is used between any two continuous columns. It ranges between $[-1,+1]$. Similarly, the \textit{Theil uncertainty coefficient} is used to measure the correlation between any two categorical features. It ranges between $[0,1]$. And the \textit{correlation ratio}  between categorical and continuous columns is used. It also ranges between $[0,1]$. Note that the dython\footnote{\url{http://shakedzy.xyz/dython/modules/nominal/\#compute\_associations}} library is used to compute these metrics. Finally, the difference between the correlation matrices for real and synthetic datasets is computed and abbreviated as \textbf{Diff. Corr.}. 
In the neural network partition experiment with two clients, to examine the preservation of column correlations within each client and between two clients, we first separately calculate the 
% difference of correlation matrices separately
\textbf{Diff. Corr.} 
for each client's data, and then name the average results \textbf{Avg-client}, we also report the difference of correlation matrices between both clients' data, the result is named \textbf{Across-client}.
% we report the difference in correlation matrices separately for data within  two clients and data across two clients, namely \textbf{Avg-client} and \textbf{Across-client}. 
Assuming clients A and B, with their local real data $RD_A$ and $RD_B$, respectively, produce synthetic data $SD_A$ and $SD_B$. {Avg-client} averages the \textbf{Diff. Corr.} of \{$RD_A$, $SD_A$\} and \textbf{Diff. Corr.} of \{$RD_B$, $SD_B$\}. {Across-client} calculates the $\mathit{l}^2$-norm of difference between the correlation matrix of \{$RD_A$, $RD_B$\} (i.e., pair-wise correlation between columns in $RD_A$ and $RD_B$) and the correlation matrix of \{$SD_A$, $SD_B$\}. These two metrics are designed to demonstrate the ability of \algo to capture intra- and inter-dependencies of training data columns within and across clients.
\subsection{Result Analysis}
% \lc{Results are sufficient. But, some more experiments shall be here to validate some of design choices. More sensitivity analysis is needed. }
The goal of the experiments is to determine the optimal configuration of \algo for synthesizing high-fidelity tabular data. To this end, we aim to answer three research questions: 
(1) What is the optimal partition of generator and discriminator neural networks between the server and clients?
(2) How does \algo respond to variations in number of data features across clients.
(3) How does \algo adapt to changes in the number of clients participating in the system.
% (1) Given that all clients have the same number of data columns, what is the best partitioning of generator and discriminator neural networks between the server and clients? 
% %\zz{add new experiments}
% (2) Given the optimal partitioning of generator and discriminator neural networks,  how will \algo react to different data distribution across clients.
% (3) Given the optimal partitioning of generator and discriminator neural networks, how will \algo react to different number of clients.
% what is the best data distribution across the clients? 
To answer these questions, we have designed three experiments. 
% Since our use case involves vertical federated learning, the number of clients is limited compared to horizontal federated learning scenarios. Therefore, all experiments involve a system with two clients. 
The centralized baseline serves as the basic tabular GAN algorithm for \algo to distribute and train in VFL. 
{In practice, the number of clients in a VFL scenario is constrained by the number of features in the tabular dataset. Therefore, following the VFL experiments setup in \cite{luo2021feature,fu2022label,fu2022blindfl}, our first two experiments are performed in a two-client setting. The \algo architecture can be readily expanded to accommodate more clients. We assess the influence of the number of clients in Section \ref{sssec:client_num}.}

\subsubsection{Neural Network Partition}
\label{sssec:nn_partition}
\begin{figure*}[t]
	\begin{center}
		% \subfloat[Difference in ML Utility]{
		% 	\includegraphics[width=0.32\textwidth]{figures/network_partition-1.png}
		% 	\label{fig:nn_partition_utility}
		% }
		% \subfloat[Statistical Similarity Difference]{
		% 	\includegraphics[width=0.3142\textwidth]{figures/network_partition-2.png}
		% 	\label{fig:nn_partition_ss}
		% }
		% \subfloat[Correlation Difference]{
		% 	\includegraphics[width=0.314\textwidth]{figures/network_partition-3.png}
		% 	\label{fig:nn_partition_cd}
		% }
        \includegraphics[width=0.95\textwidth]{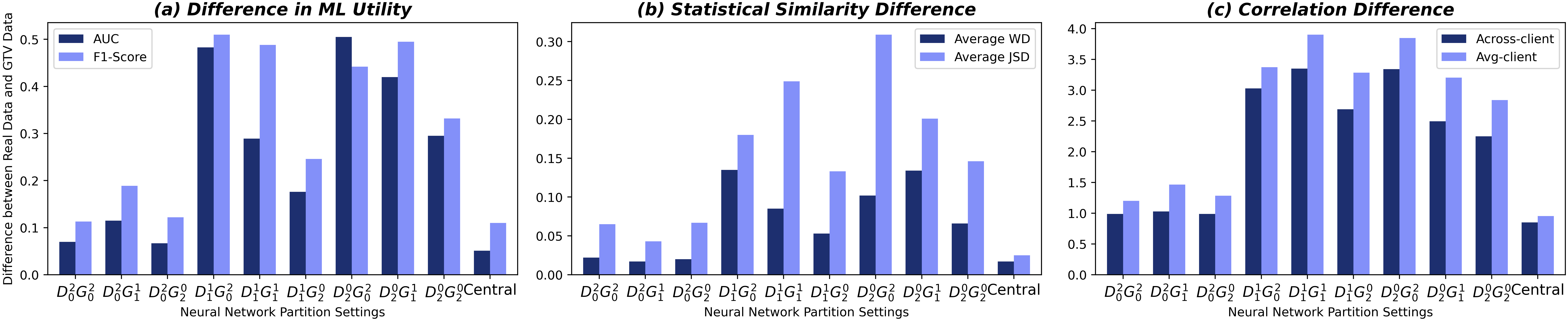}
  \vspace{-0.7em}
		\caption{ Neural network partition results: the difference between real and \algo data. A lower value indicates higher quality.}
  %\lc{do these results make sense? I would expect ML utility shows opposite trend than the similarity}} 
		\label{fig:nn_partition_result}
 	\end{center}
\vspace{-1em}
\end{figure*}
% \begin{figure*}[t]
% 	\begin{center}
% 		\subfloat[ML Utility]{
% 			\includegraphics[width=0.33\textwidth]{figures/network_partition-1.png}
% 			\label{fig:nn_partition_utility}
% 		}
% 		\subfloat[Statistical Similarity]{
% 			\includegraphics[width=0.325\textwidth]{figures/network_partition-2.png}
% 			\label{fig:nn_partition_ss}
% 		}
% 		\subfloat[Correlation Difference]{
% 			\includegraphics[width=0.32\textwidth]{figures/network_partition-3.png}
% 			\label{fig:nn_partition_cd}
% 		}
% 		\caption{Neural Network Partition Results} 
% 		\label{fig:nn_partition_result}
%  	\end{center}
% % \vspace{-2em}
% \end{figure*}

For this experiment, we evenly split the training datasets for two clients (or one client may have one more column if the total number of columns is odd), and the column orders are preserved as they were downloaded. As illustrated in Fig.~\ref{fig:partition_structure}, we propose three ways to partition the neural networks of both the generator and discriminator across the server and clients. Both the generator and discriminator have two network blocks (same as in centralized baseline), so we consider three divisions for each of them: (i) all blocks in the server, (ii) one block in the server and one block in each client, (iii) all blocks in the client. {This results in a total of nine combinations. The partition notions of $G^{n_1}_{n_2}$ and $D^{n_3}_{n_4}$ are explained in Fig.~\ref{fig:partition_structure}.}
% \lc{The following description depends very much on how network architecture was decided in our setup. We should note that very clearly. And, provide the explaination how this will look like for different number of blocks} \lc{Can a better explaination be provided here? Difficult to follow the notiations}The notation $D^{N_1}_{N_2}G^{N_3}_{N_4}$ represents a configuration where the discriminator and generator have $N_1$ FN and $N_3$ RN block(s) on the server and $N_2$ FN and $N_4$ RN block(s) on the client, respectively. $N_1 + N_2 = N_3 + N_4 = 2$ for all combinations. 
The sum of the output dimension of partitioned RN/FN block is the same as in their centralized scenario, i.e., 256. The output dimension of each partitioned block is calculated based on the ratio vector $P_r$ for each client.

Figure~\ref{fig:nn_partition_result} shows the neural network partition results where all results are averaged over five datasets. In all three subfigures, it is clear that the centralized method performs best on all metrics, as expected. Of the other partition methods, $D^{2}_{0}G^{2}_{0}$, $D^{2}_{0}G^{1}_{1}$, and $D^{2}_{0}G^{0}_{2}$ stand out as they consistently outperform the other six configurations on all metrics. These three partition methods have all the FN blocks on the server side. Among these three, 
%they all perform similarly for correlation difference. 
$D^{2}_{0}G^{1}_{1}$ performs slightly better in terms of statistical similarity and slightly worse in correlation difference, but clearly worse in terms of ML utility compared to the other two. $D^{2}_{0}G^{2}_{0}$ and $D^{2}_{0}G^{0}_{2}$ show similar results across all evaluation metrics and consistently outperform all other configurations in terms of ML utility. $D^{2}_{0}G^{2}_{0}$ achieves F1-score only 2.7\% lower to the centralized method. Above result suggests that having a sufficiently large discriminator on the server side is crucial to the success of \algo, and that \algo can achieve performance similar to the centralized method if the discriminator and generator are properly positioned. 

Based on the overall performance, we consider $D^{2}_{0}G^{2}_{0}$ and $D^{2}_{0}G^{0}_{2}$ to be the two best configurations for \algo. To choose between these two configurations, we must consider the computation and communication overhead of the system. In terms of communication overhead, their only difference is the size of the intermediate logits transferred between the server and clients for generator, which can be controlled by the FC layer before logits are sent from the server to the client. In current setting, $D^{2}_{0}G^{2}_{0}$ has higher communication overhead; In terms of computation overhead, if the server is powerful, placing both the generator and discriminator on the server can speed up the training process. However, many computations are now performed on the cloud, there is a cost associated with using the server. It may be more cost-effective to distribute calculations to the clients. Additionally, as the number of clients increases, the generator can become larger, making it more expensive to place the entire generator on the server side. Therefore, when $D^{2}_{0}G^{2}_{0}$ and $D^{2}_{0}G^{0}_{2}$ achieve the similar result, $D^{2}_{0}G^{0}_{2}$ is preferred. 

\subsubsection{Training Data Partition}
% \zz{table 2 can be removed, also in the text, mention the difference across datasets.}

\label{exp:training_data_partition}
\begin{figure}[t]
    \centering
    \includegraphics[width=0.75\linewidth]{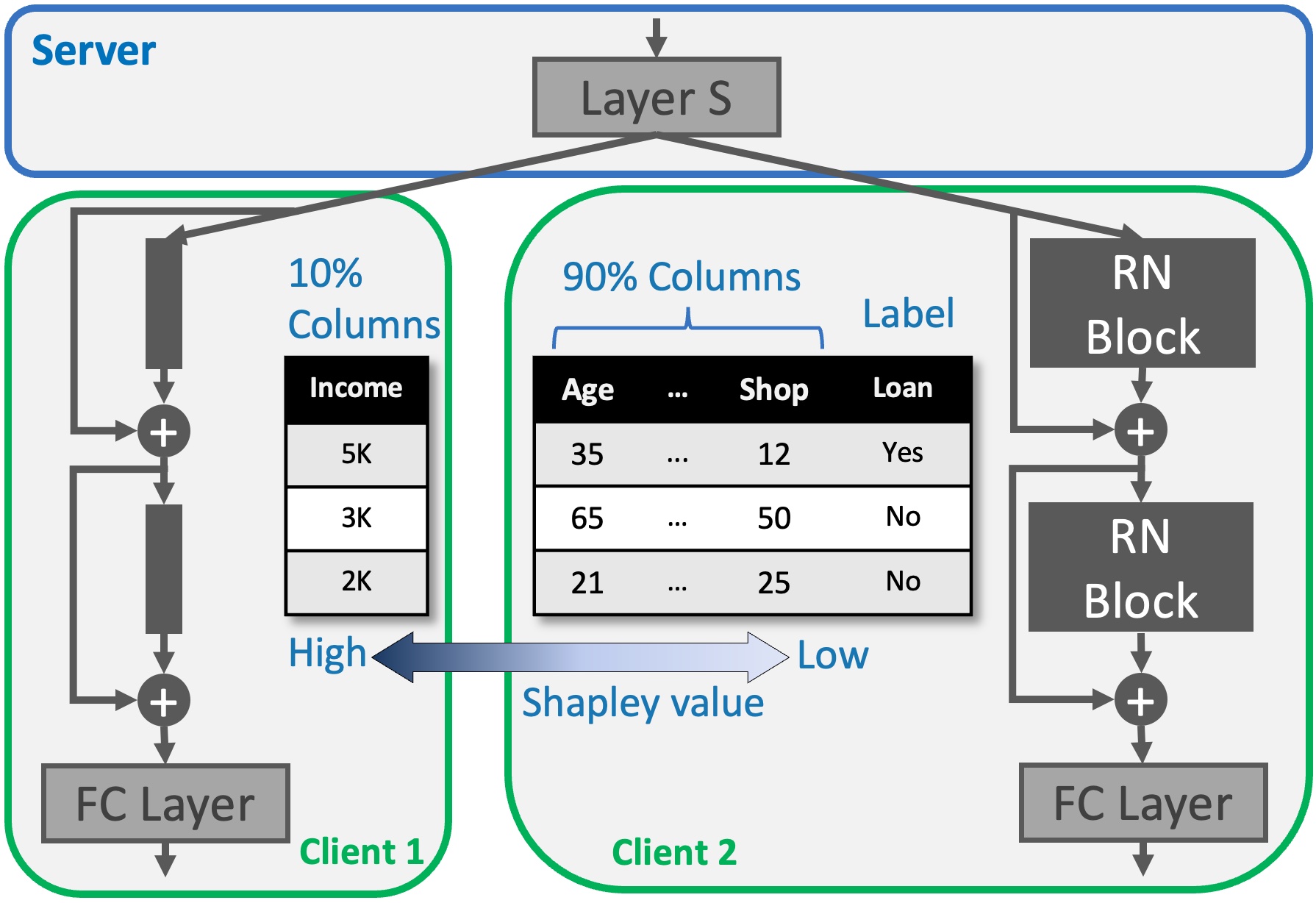}
    \vspace{-0.5em}
    \caption{An example of the $G^{0}_{2}$ generator partition in 1090 data partition experiment.}
    \label{fig:g02-1090}
   \vspace{-1.5em}
\end{figure}

\begin{figure*}[t]
    \centering
    \includegraphics[width=0.95\linewidth]{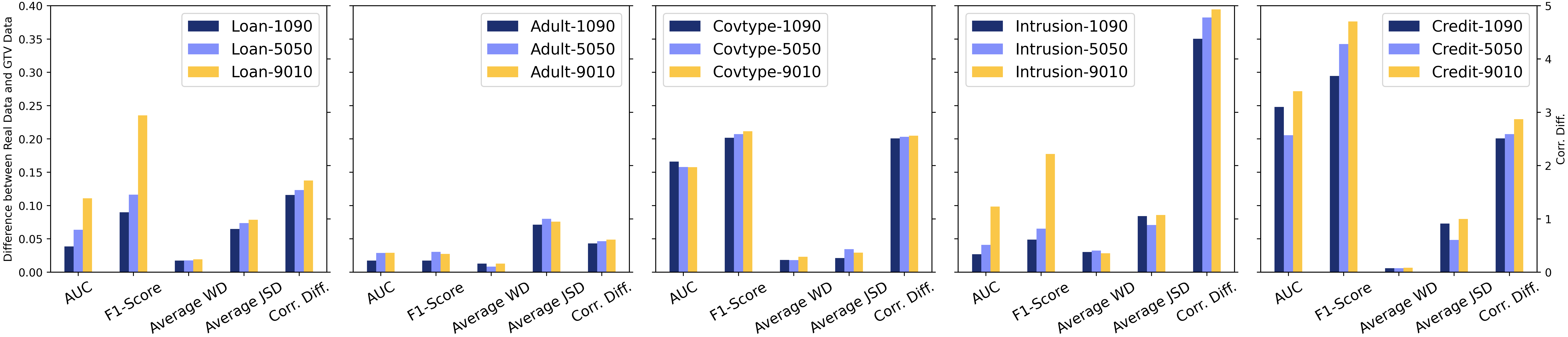}
    %\vspace{-1em}
    \vspace{-1.em}
    \caption{Data partition results with $D^{2}_{0}G^{0}_{2}$: the difference between real and \algo data. Corr.Diff. (right axis).}
    \label{fig:datapartition_result}
   \vspace{-0.7em}
\end{figure*}

\begin{figure*}[t]
    \centering
    \includegraphics[width=0.95\linewidth]{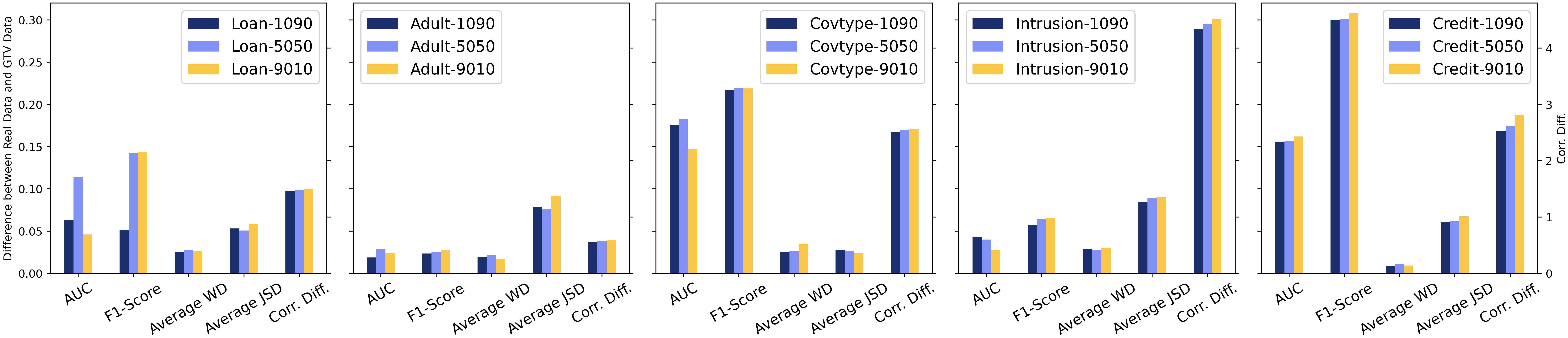}
    \vspace{-1.em}
    % \vspace{-0.5em}
    \caption{Data partition results with $D^{2}_{0}G^{2}_{0}$: the difference between real and \algo data. Corr.Diff. (right axis)}
    \label{fig:datapartition_result_all_server_side}
   \vspace{-1em}
\end{figure*}

Experiment in Sec.~\ref{sssec:nn_partition} suggests that $D^{2}_{0}G^{2}_{0}$ and $D^{2}_{0}G^{0}_{2}$ are two optimal configurations for \algo when data features are randomly and evenly distributed among two clients. In this experiment, we examine the impact of data partition on these setups of \algo. 
% We use the feature importance value evaluated by Shapley value that we calculated in Sec.~\ref{ssec:motivation}. 
For each dataset, we use the Shapley value~\cite{lundberg2017unified} to evaluate the importance of each feature in predicting the target column using a MLP (Multi-layer Perceptron) model with one hidden layer containing 100 neurons. 
% we first use Shapley value~\cite{lundberg2017unified} to measure the importance of each dataset's features for predicting the target column using a MLP model with one hidden layer containing 100 neurons. 
We sort the features importance and consider three different feature divisions: (i) 1090: the 10\% most important features and the remaining 90\%, (ii) 5050: 50\% most important features and 50\% remaining features, and (iii) 9010: 90\% most important features and 10\% remaining features. For each division, we assign the two parts to two clients. The target column is always located on the client WITHOUT the most important features. 
The intuition behind these designs is that it is easier to learn the correlations among features within a single client than it is to learn correlations among features distributed across multiple clients. Fig.~\ref{fig:g02-1090} shows an example 1090 data partition for $G^{0}_{2}$.  It is worth noting that when partitioning the data, we also split the RN block proportionally among the clients based on the ratio vector $P_r$.
%proportion of the number of features in each client to the total number of features. 
But the sum of the output dimension of the partitioned block is still the same as the centralized baseline, i.e., 256.
% \begin{figure*}[t]
%     \centering
%     \includegraphics[width=1\linewidth]{figures/data_partition.png}
%     %\vspace{-1em}
%     \caption{Data Partition Results with $D^{2}_{0}G^{0}_{2}$.}
%     \label{fig:datapartition_result}
%    %\vspace{-1em}
% \end{figure*}

% Fig.~\ref{fig:datapartition_result} summarizes the results for all five datasets and we see different performances on different datasets. We first focus on the ML utility metrics. For Adult and Covtype datasets, the data partition has less impact on the final synthetic data. But for Loan, Intrusion, and Credit datasets, the influence is much higher, the 9010 partition makes their results much worse than the other two configurations. Being the impact big or small, we observe on all datasets that there is a trend that the result of 1090 is better than 5050, and 5050 is better than 9010. When we compare Fig.~\ref{fig:motivation} and Fig.~\ref{fig:datapartition_result}, we can find that the three most sensitive datasets to data partition are also the three datasets that achieve best F1-score in Fig.~\ref{fig:datapartition_result}. The reason can be that in these three datasets, there are several important features for predicting target column. As they are excluded step-by-step from the label holder, that influences the ability of \algo to fully capture the correlations between these features and the label.

Fig.~\ref{fig:datapartition_result} shows the results for  $D^{2}_{0}G^{0}_{2}$  on five datasets. Focusing on the ML utility metrics, we see that data partition has a smaller impact on the final synthetic data for the Adult and Covtype datasets, but a much larger impact on the Loan, Intrusion, and Credit datasets, where the 9010 partition leads to significantly worse results compared to the other two configurations. Regardless of the size of the impact, we observe on almost all datasets that the 1090 partition consistently performs better than the 5050 partition, which in turn performs better than the 9010 partition. 
% When comparing Fig.~\ref{fig:motivation} and Fig.~\ref{fig:datapartition_result}, we find that the three datasets most sensitive to data partition are also the ones that achieve the best F1-scores in Fig.~\ref{fig:datapartition_result}. 
This because each dataset contains several important features for predicting the target column, and as these features are excluded step-by-step from the label holder, it reduces \algo's ability to fully capture the correlations between these features and the label.
The results for statistical similarity are similar across all datasets and configurations. The results of the Diff.Coff. align with the ML utility results, which is expected since the preservation of column correlations is inherently linked to ML utility.
%{Diff.Coff. results are inline with the results of ML utility, which is intuitive because the preservation of column correlations should be related to the ML utility.}
% Since the Diff. Coff. is on a different scale than the other metrics, it is not included in the same figure. Instead, it is shown in Tab.~\ref{table:corrdiff}. This result supports the finding from the ML utility metric, where the 1090 partition consistently demonstrates better feature correlation than the other two configurations. This validates our hypothesis that a larger number of features on the label holder allows \algo to more accurately capture correlations between features and label, resulting in the improvement in ML utility.

% For the results on statistical similarity, the results for all the datasets are similar among the three configurations. Since Coff. Diff. is on a different scale than other metrics, it is not illustrated within the same figure, we show this result in Tab.~\ref{table:corrdiff}. This result confirms the result for ML utility, the 1090 always achieves a better feature correlation than the other two configurations. That result confirms our hypothesis that the more features on the label holder, the better the model can capture the feature correlations, which leads to better ML utility.  

% \begin{figure*}[t]
%     \centering
%     \includegraphics[width=1\linewidth]{figures/data_partition_G20D20.png}
%     %\vspace{-1em}
%     \caption{Data Partition Results with $D^{2}_{0}G^{2}_{0}$.}
%     \label{fig:datapartition_result_all_server_side}
%    %\vspace{-1em}
% \end{figure*}

The results for $D^{2}_{0}G^{2}_{0}$ on the five datasets are shown in Figure~\ref{fig:datapartition_result_all_server_side}. In terms of ML utility, especially the F1-Score, the 1090 partition performs best on all datasets. The results on Loan dataset are still affected by the data partition, However, it is also clear that $D^{2}_{0}G^{2}_{0}$ is much less impacted compared to $D^{2}_{0}G^{0}_{2}$. The results for statistical similarity for $D^{2}_{0}G^{2}_{0}$ are similar to those for $D^{2}_{0}G^{0}_{2}$, with little impact from the different data partitions on \algo's ability to recover the column-wise distribution in the synthetic data. {The Diff. Corr. shows that the 1090 partition still outperforms the other two configurations. But comparing to $D^{2}_{0}G^{0}_{2}$, $D^{2}_{0}G^{2}_{0}$ is less affected by data partition. The reason is that the larger generator on the server side is able to better capture the column correlations across clients.}

% The Diff. Corr., shown in Table~\ref{table:corrdiff}, reveals that the 1090 partition outperforms the other two configurations, and that $D^{2}_{0}G^{2}_{0}$ is less affected by data partition. The reason that $D^{2}_{0}G^{2}_{0}$ outperforms $D^{2}_{0}G^{0}_{2}$ on ML utility and coefficient of difference for the 9010 partition is that the larger generator on the server side is able to better capture the column correlations across clients.

% Result for $D^{2}_{0}G^{2}_{0}$ on five datasets is shown in Fig.~\ref{fig:datapartition_result_all_server_side}. For the ML utility results, we can still observe that 1090 get the best results in all the datasets. But it is also clear that $D^{2}_{0}G^{2}_{0}$ shows much less impact by the data partition compared to $D^{2}_{0}G^{0}_{2}$. The statistical similarity result for $D^{2}_{0}G^{2}_{0}$ is similar as for $D^{2}_{0}G^{0}_{2}$, different data partition has little impact for \algo to recover column-wise distribution in synthetic data. Diff. Corr. in Tab.~\ref{table:corrdiff} shows that 1090 outperforms other two configurations and $D^{2}_{0}G^{2}_{0}$ is less influenced by data partition. The reason $D^{2}_{0}G^{2}_{0}$ outperforms $D^{2}_{0}G^{0}_{2}$ on ML utility and Diff. Corr. for 9010 is because the bigger server side generator can better capture the column correlations across clients. 

%And the better result on 1090 reconfirms our hypothesis that a larger number of features on the label holder improves \algo on ML utility and Diff. Corr..

Comparing the result of $D^{2}_{0}G^{0}_{2}$ and $D^{2}_{0}G^{2}_{0}$ in the data partition experiment provides additional information to help us choose between these two configurations. In neural network partition experiment, when features are evenly and randomly distributed to clients, $D^{2}_{0}G^{0}_{2}$ and $D^{2}_{0}G^{2}_{0}$ achieve similar result, leading us to prefer $D^{2}_{0}G^{0}_{2}$ due to its lower cost and better scalability. However, when the data partition is extremely imbalanced, especially when the label holder contains significantly fewer features than other clients, $D^{2}_{0}G^{2}_{0}$ becomes the preferred configuration. {One possible solution to increase the performance of $D^{2}_{0}G^{0}_{2}$ is to increase the size of neural network on the client who contains fewer features, this left for future work.}

One might argue that when organizing a vertical federated learning group, the data partition is fixed and determined by the clients. However, when clients join vertical federated learning, it is possible for different clients to have overlapping features, and in such cases, we have the option to distribute more features to the label holder to improve ML utility for the synthetic data. This is because we only need one of the overlapping features to be present in the system. For datasets without a target column, we can try to balance the feature distribution among clients to avoid artificially creating imbalances that could degrade the training of \algo.

% \begin{figure}[h]
%     \centering
%     \includegraphics[width=0.99\linewidth]{figures/D20G20_num_clients.png}
%     %\vspace{-1em}
%     \caption{Performance of $D^{2}_{0}G^{2}_{0}$ over various number of clients}
%     \label{fig:d20g20_num_clients}
%    %\vspace{-1em}
% \end{figure}

% \begin{figure}[h]
%     \centering
%     \includegraphics[width=0.99\linewidth]{figures/D20G02_num_clients.png}
%     %\vspace{-1em}
%     \caption{Performance of $D^{2}_{0}G^{0}_{2}$ over various number of clients}
%     \label{fig:d20g02_num_clients}
%    %\vspace{-1em}
% \end{figure}

% table 2
% \begin{table}[t]
% \centering
% \caption{Result of Corr. Diff. on Data Partition}
% \vspace{-1em}
% \resizebox{1\columnwidth}{!}{
% \begin{tabular}{c c c c c c}
% \toprule
% \textbf{Partition-distribution} &\textbf{Loan} &\textbf{Adult} &\textbf{Covtype}&\textbf{Intrusion}&\textbf{Credit} \\
% \hline
% $D^{2}_{0}G^{0}_{2}$-1090 &{1.45}&{0.54}&{2.51}&{4.38}&{2.51}\\[0.35em]
% $D^{2}_{0}G^{0}_{2}$-5050 & 1.54 & 0.58 & 2.54 & 4.78 & {2.59}\\[0.35em]
% $D^{2}_{0}G^{0}_{2}$-9010 &1.72&{0.61}&2.56&4.93&2.87\\[0.35em]
% \hline
% $D^{2}_{0}G^{2}_{0}$-1090 &1.46&{0.55}&2.51&4.34&2.53\\[0.35em]
% $D^{2}_{0}G^{2}_{0}$-5050 &1.48&{0.58}&2.55&4.43&2.61\\[0.35em]
% $D^{2}_{0}G^{2}_{0}$-9010 &1.50&{0.59}&2.56&4.51&2.81\\[0.35em]
% \bottomrule
% \end{tabular}
% }
% \label{table:corrdiff}
% \vspace{-1em}
% \end{table} 

\subsubsection{Client Number Variation}
\label{sssec:client_num}
% \zz{Fig 12 and 13 now includes Corr. Diff., so Table 3 may be moved to the appendix.}
For this experiment, we study the impact of number of clients on \algo.  We randomly and evenly distribute data columns into 2, 3, 4 and 5 clients on $D^{2}_{0}G^{2}_{0}$ and $D^{2}_{0}G^{0}_{2}$ configurations. For each dataset, the number of columns is fixed, the more clients in the system, the less data columns for each client. To address the potential performance degradation caused by an increasing number of clients in the system, we introduced two settings for the generator: (1) \textit{default} and (2) \textit{enlarged}. In the \textit{default} setting, for $D^{2}_{0}G^{0}_{2}$, as the number of clients increases, we simply create more partitions of the original RN block. However, the sum of the output dimension of the divided RN block remains constant, at 256, as in the centralized case. The output dimension of all RN blocks is also 256 for $D^{2}_{0}G^{2}_{0}$. \textit{Default} setting ensures, no matter how many clients join the system, the communication overhead between server and clients remains the same for $D^{2}_{0}G^{2}_{0}$ and $D^{2}_{0}G^{0}_{2}$, respectively; For the \textit{enlarged} setting, we increase the output dimension of RN block to 768 ($3\times 256$) for both $D^{2}_{0}G^{2}_{0}$ and $D^{2}_{0}G^{0}_{2}$, We will explain the reason for choosing this number with the results later.

% \begin{figure}[t]
%     \centering
%     \includegraphics[width=0.99\linewidth]{figures/D20G20_num_clients_new.png}
%     %\vspace{-1em}
%     \caption{Performance of $D^{2}_{0}G^{2}_{0}$ over various number of clients: default generator (firm line), enlarged generator (dashed line)}
%     \label{fig:d20g20_num_clients}
%     \vspace{-1em}
% \end{figure}

% \begin{figure}[t]
%     \centering
%     \includegraphics[width=0.99\linewidth]{figures/D20G02_num_clients_new.png}
%     %\vspace{-1em}
%     \caption{Performance of $D^{2}_{0}G^{0}_{2}$ over various number of clients: default generator (firm line), enlarged generator (dashed line)}
%     \label{fig:d20g02_num_clients}
%    \vspace{-1em}
% \end{figure}

\begin{figure}[t]
    \centering
    \includegraphics[width=0.92\linewidth]{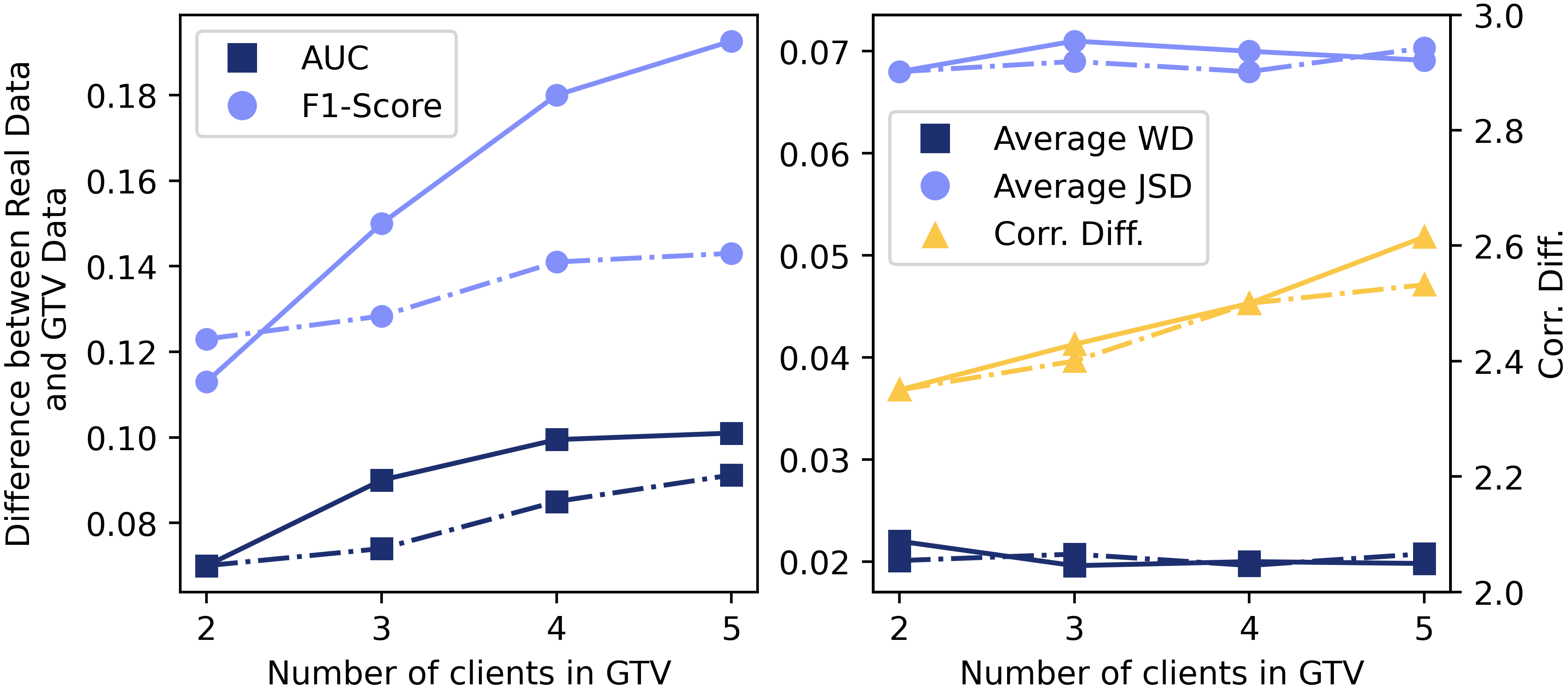}
    \vspace{-0.7em}
    \caption{$D^{2}_{0}G^{2}_{0}$ setting, the diff. between real and \algo data over varied numbers of clients: default generator (firm line), enlarged generator (dashed line), Corr.Diff. (right axis).}
    \label{fig:d20g20_num_clients}
    \vspace{-1.2em}
\end{figure}

\begin{figure}[t]
    \centering
    \includegraphics[width=0.92\linewidth]{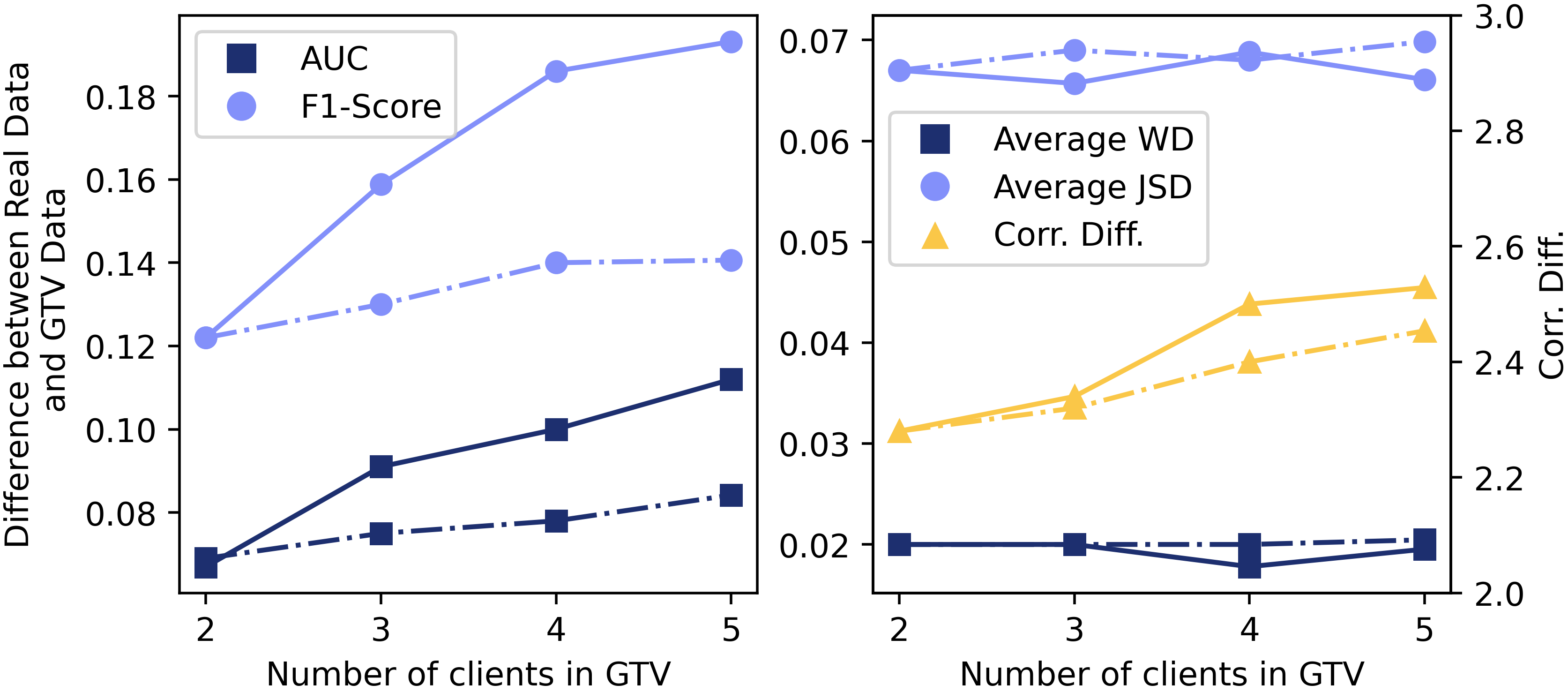}
    \vspace{-0.7em}
    \caption{$D^{2}_{0}G^{0}_{2}$ setting, the diff. between real and \algo data over varied numbers of clients: default generator (firm line), enlarged generator (dashed line), Corr.Diff. (right axis).}
    \label{fig:d20g02_num_clients}
   \vspace{-1.2em}
\end{figure}

Fig.~\ref{fig:d20g20_num_clients} and ~\ref{fig:d20g02_num_clients} present the results on five metrics, with solid lines representing the results of the \textit{default} setting and dashed lines representing the results of the \textit{enlarged} setting. All results are averaged over five datasets. With \textit{default} setting, the results show that with more clients in the system, i.e., each client contains less features, the ML utility of synthetic data becomes slightly worse. For example, as the number of clients increases from 2 to 5, $D^{2}_{0}G^{2}_{0}$ and $D^{2}_{0}G^{0}_{2}$ increase their F1-Score from 0.11 and 0.12 to 0.19, respectively; With \textit{enlarged} setting, ML utility also decreases with increasing number of clients, but to a significantly lesser extent; The results of the Average WD and Average JSD indicate that the statistical similarity on both generator settings for both $D^{2}_{0}G^{2}_{0}$ and $D^{2}_{0}G^{0}_{2}$ remains largely unchanged regardless of the number of clients. {Again, Coff.Diff. results are inline with ML utility. The result becomes worse with increasing number of clients for both settings. In most of scenarios, $D^{2}_{0}G^{0}_{2}$ slight outperforms $D^{2}_{0}G^{2}_{0}$.}

%Due to scale difference, we show the Coff.Diff. in Tab.~\ref{table:corrdiff_variation_clients}, we can see the trend that we observed in the ML utility, the result becomes worse with increasing number of clients for both settings on generator for $D^{2}_{0}G^{2}_{0}$ and $D^{2}_{0}G^{0}_{2}$. In most of scenarios, $D^{2}_{0}G^{0}_{2}$ slight outperforms $D^{2}_{0}G^{2}_{0}$. 
Therefore, in that case, $D^{2}_{0}G^{0}_{2}$ is still the preferred configuration. 
The results for two settings on generator reveal that the \textit{enlarged} setting has less variation as the number of clients increases.
% , which aligns with the ML utility outcome. 
However, when there are only two clients in the system, the \textit{enlarged} setting performs worse on the Loan dataset. 
% The output dimension of the enlarged setting is determined through trial and error. It is important to consider the trade-off between the size of the neural network and the speed of model convergence. Increasing the size of the neural network does not always lead to improved results.
% As the number of clients increases, one may be inclined to increase the neural network size. However, the Loan dataset only contains 5000 data instances, and increasing the neural network size resulted in slower convergence for the generator in the enlarged setting. 
The output dimension of the \textit{enlarged} setting is determined through trial and error, highlighting the importance of considering the trade-off between the size of the neural network and the model convergence. It shows that increasing the size of the neural network does not always lead to improved results.
Given that the Loan dataset only contains 5000 data instances, increasing the neural network size resulted in slower convergence for the generator in the \textit{enlarged} setting.
% To ensure a fair comparison, we always sample synthetic data from the 300th epoch. As a result, the performance of the \textit{enlarged} setting is inferior. 
Optimal neural network size exploration for \algo under different dataset and different number of clients is left for future work.

\subsubsection{{Membership Inference Attack on \algo with Differential Privacy}}
\label{sss:dp_gtv}
% Since the goal of \algo is to publish the synthetic data, there are potential risk that the synthetic data can cause privacy leakage of original data. Therefore, we test differential privacy (DP) in our system in order to protect original data. Specifically, we adopts DP-SGP~\cite{abadi2016deep} on discriminator training since discriminator needs real data to train itself. Fig.~\ref{fig:gtv-dp} shows the result on $D^{2}_{0}G^{2}_{0}$ setting with random and evenly split features on two clients, same as in experiment in Sec.~\ref{sssec:nn_partition}. We fix the privacy budget $\epsilon$ to three commonly used value 1 5 and 10~\cite{dpvalue} and compare their synthetic data quality with the same setting without DP. Higher privacy budget means lower privacy guarantee. Results averaged on 5 datasets show that DP largely reduces the synthetic data quality in all metrics. DP makes it difficult to infer the original data distribution according to synthetic data (i.e., higher Average JSD and Average WD.). Larger privacy budget improves synthetic data quality，but the result is still far from the scenario without DP. 
{As we discussed in Section \ref{ssec:security_analysis}, \algo can be vulnerable to the full black-box MIA \cite{hilprecht2019monte, chen2020gan, hu2021tablegan}. In our experiments, we conduct \cite{chen2020gan}'s full black-box MIA on \algo, and evaluate the performance of this attack on the five datasets. Moreover, we introduce Differential Privacy (DP) into \algo as it is considered the most effective defense mechanism against MIA \cite{xie2018differentially,chen2020gs,torfi2020differentially}. Specifically, we use the differential private (DP) stochastic gradient descent(SGD) ~\cite{abadi2016deep} for training the discriminator (both client and server side) since only its training requires real data.} 
{The algorithm can be summarized into a two-step process.  Firstly, the per-sample gradient computed during each training iteration is subject to clipping by its $L2$ norm using a predetermined threshold. Secondly, calibrated random noise is deliberately incorporated into the gradient to introduce stochasticity, thereby safeguarding privacy during the training process.}

\begin{figure}[t]
    \centering
    \includegraphics[width=0.70\linewidth]{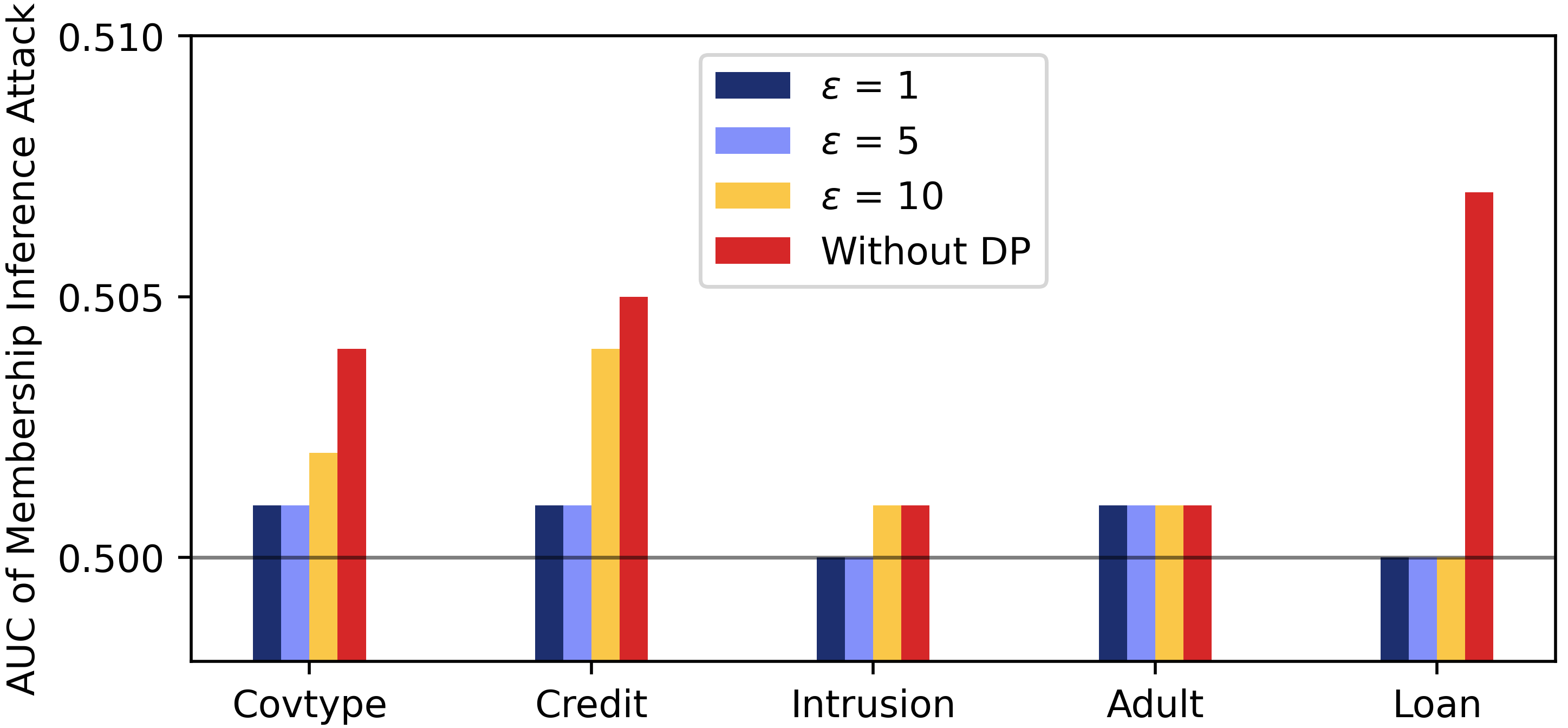}
    \vspace{-0.5em}
    \caption{The result of Membership Inference Attack (full black-box) \cite{chen2020gan} on \algo, under different DP settings.}
    \label{fig:gtv-dp-mia}
   \vspace{-1.5em}
\end{figure}

{The results of the MIA conducted on all five datasets are presented in Fig.~\ref{fig:gtv-dp-mia}. In DP-SGD, the privacy budget $\epsilon$ determines the level of privacy protection. Smaller $\epsilon$ values offer stronger privacy guarantees but introduce larger amounts of noise into the model. It is evident that even without DP, the MIA is not highly effective, as a random guess model would achieve an AUC of 0.5. The AUC for the loan dataset is slightly higher than that for the other datasets because it contains fewer data records, which makes it more vulnerable to attack \cite{chen2020gan}. Nonetheless, despite the effect is weak, we observe that for the Covtype and Credit dataset, adopting DP reduces the AUC of the MIA.}

\begin{figure}[t]
    \centering
    \includegraphics[width=0.75\linewidth]{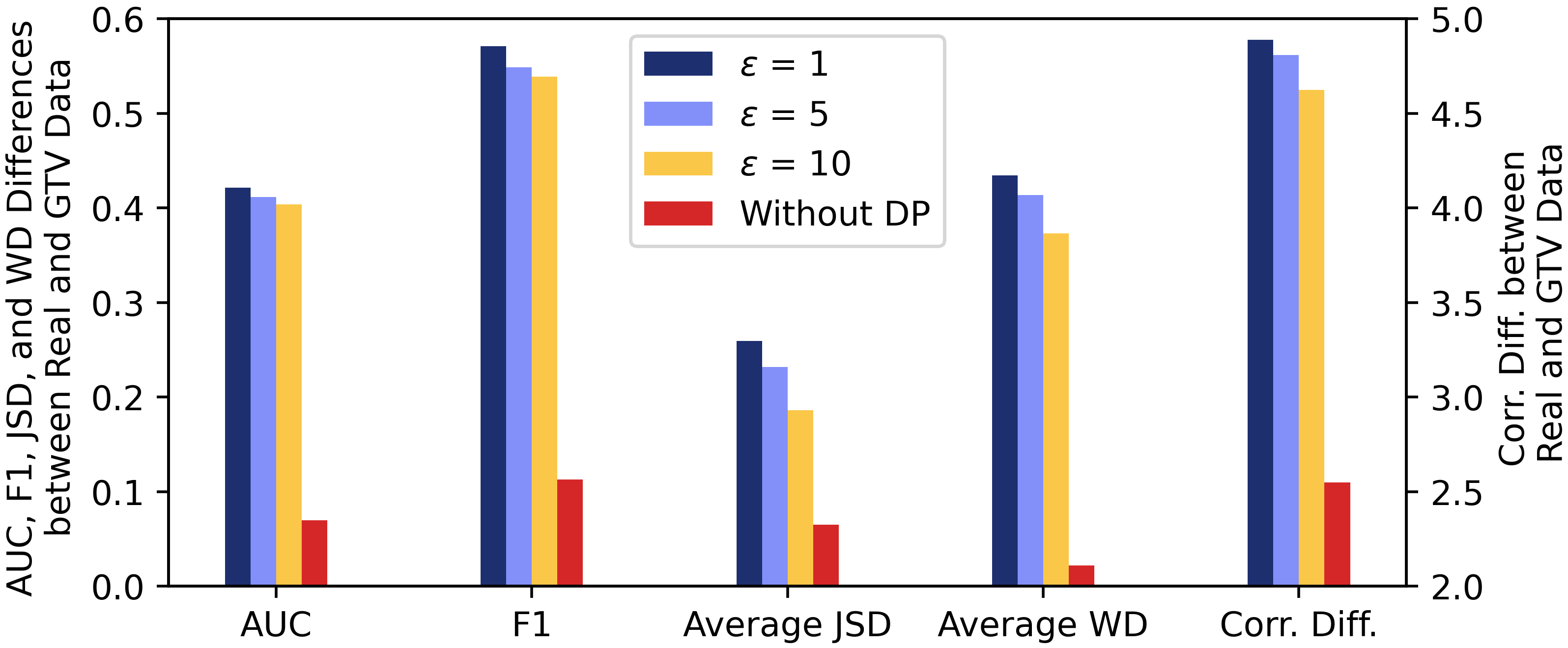}
    \vspace{-0.5em}
    \caption{Under $D^{2}_{0}G^{2}_{0}-5050$ setting, The impact of Differential Privacy on \algo performance. {Corr.Diff. (right axis)}}
    \label{fig:gtv-dp}
   \vspace{-1.2em}
\end{figure}

{DP affects the utility of model due to the addition of random noise during the training process. As for \algo, we  evaluate the impact of DP on the synthetic data quality. We conduct experiments on the $D^{2}_{0}G^{2}_{0}$ setting with random and evenly split features on two clients, same as setting in Sec.~\ref{sssec:nn_partition} where $D^{2}_{0}G^{2}_{0}$ performs among the best. We fix the privacy budget $\epsilon$ to three commonly used values (i.e., 1, 5, and 10~\cite{dpvalue}) and compare the synthetic data quality with and without DP. Our results in Fig.~\ref{fig:gtv-dp}, averaged over five datasets, indicate that DP significantly reduces the quality of the synthetic data in all metrics. Specifically, the average JSD and average WD between the original and synthetic data increase with DP. These results suggest that it becomes more challenging to infer the original data distribution from the synthetic data. Increasing the privacy budget can enhance the synthetic data quality, but the improvement is still inadequate compared to the scenario without DP. }

\textbf{Take Aways}
The experiments show that a key factor for a successful \algo is having a sufficiently large discriminator model on the server side. When the number of data columns is evenly distributed among two clients in \algo, both $D^{2}_{0}G^{0}_{2}$ and $D^{2}_{0}G^{2}_{0}$ exhibit similar performance and are comparable to the centralized algorithm. In cases where these two configurations yield comparable results, $D^{2}_{0}G^{0}_{2}$, i.e., distributes most of the generator network to client, is preferable due to its superior scalability and cost-effectiveness. However, when data is unevenly distributed among clients, then $D^{2}_{0}G^{2}_{0}$, i.e., distributes most of the generator network to server, is preferred. If control over the distribution of data columns among clients is possible, e.g., excluding certain overlapping data columns in certain clients, it is recommended to balance the number of columns among the clients in the system. With a fixed total number of features in the system, an increasing number of clients in \algo reduces the quality of synthetic data. Strategically increasing the generator model size can effectively counter this performance degradation. For the security analysis, despite the limited effectiveness of DP in countering MIA, its adoption in \algo significantly diminishes the quality of synthetic data, leading us to discourage the use of strong DP measures.

 \section{Related Work}
%\lc{move this to the second last section}
\subsection{Vertical Federated Learning} 
% \wh{Related literature around VFL}

While Horizontal Federated Learning (HFL) has been thoroughly studied \cite{mcmahan2017communication, yang2019federated}, Vertical Federated Learning (VFL) starts to gain more attention more recently~\cite{hu2019fdml,wu2020privacy,liu2020federated,cheng2021secureboost,fu2021vf2boost,fu2022blindfl,fu2022towards}. Among them, \cite{hu2019fdml} assumes that the label holder client's labels can be directly shared with others, which directly violates privacy. SecureBoost is proposed in \cite{cheng2021secureboost} to use a lossless approach for training Gradient-Boosted Decision Trees (GBDT) in VFL scenario. VF$^{2}$Boost \cite{fu2021vf2boost} focuses on improving the efficiency of the GBDT algorithm under VFL setting by adopting a concurrent training protocol to reduce the idle periods. Liu et al. \cite{liu2020federated} achieve VFL Random Forest (RF) training based on the assumption of a third-party trusted server. In \cite{wu2020privacy}, the authors consider the use of Classification and Regression Trees (CART) in a VFL setting, which does not require the involvement of a third party but has the drawback of exponential complexity as the tree becomes deeper. Homomorphic Encryption (HE) and Secrete Sharing technologies are comprehensively applied in BlindFL \cite{fu2022blindfl} to protect VFL from most privacy attacks. An efficient VFL training framework with cached-enable local updates is proposed in \cite{fu2022towards} to address the communication bottleneck across multiple participants. Li et al. \cite{li2021federated} address the matrix factorization problem in VFL for training recommendation models. Though not designed for VFL, He et al. \cite{he2020transnet} propose TransNet to encrypt vertically partitioned data before sending it to the neural network, whereas the bottom model in VFL offers similar functionality. FedDA~\cite{zhang2022data} trains separate GANs in each participant under VFL to achieve data augmentation. However, their study is centered on image generation, which is not a natural use case for VFL. Furthermore, their study 
% is limited to only two clients and 
does not incorporate the use of a conditional vector. \algo trains only one GAN, adopts conditional vector, and supports multiple clients.
%\lc{Add a sentence to stress the benefit of GTV}

\subsection{Privacy Preserving Techniques}
{Differential Privacy (DP)\cite{dwork2014algorithmic} has been extensively studied for GANs \cite{xie2018differentially,chen2020gs,torfi2020differentially} and in HFL scenarios \cite{sun2019can,mcmahan2017learning,geyer2017differentially,naseri2020local} to provide provable privacy guarantees. In VFL with graph data processing, DP can be achieved by injecting noise into the sample representation \cite{qiu2022your}.  Local Differential Privacy (LDP) has also been applied to VFL tree boosting models in a recent study \cite{li2022opboost}. However, there is currently no clear guidance on how to apply DP in VFL when dealing with both GANs and tabular data, i.e., the scenario of \algo. Moreover, DP's utility is significantly limited due to the negative impact of noise injection on model performance, leading most VFL studies to deprecate its use \cite{fu2022blindfl, fu2021vf2boost, wu2020privacy, liu2020federated, he2020transnet, zhang2022data}.
% We do not apply DP in \algo as this loss of model accuracy is contrary to \algo’s purpose of using combined datasets to improve synthetic data utility. 
Another solution to enhance privacy in \algo is through the use of multi-party computation (MPC) techniques including homomorphic encryption (HE) and secret sharing (SS) \cite{mohassel2018aby3,fu2022blindfl, liu2020federated, he2020transnet}. 
%MPC allows multiple clients to compute the feedforward and backpropagation results based on encrypted data. 
BlindFL\cite{fu2022blindfl} presents the federated source layer based on the HE and SS techniques to achieve promising privacy guarantees without affecting VFL model accuracy. 
Although this approach is compatible with \algo, its high communication cost makes it less desirable compared to the privacy guarantees already provided by the \algo architecture.}

\subsection{GANs for Tabular Data}
GANs are initially known for its success in image synthesis. However, with its ample application scenarios in areas such as medicine~\cite{medgan} and finance~\cite{finance}, research on GAN for tabular data synthesis also gains attention.
% \tablegan~\cite{tablegan} implements an auxiliary classification model along with discriminator training to enhance column dependency in the synthetic data.  \ctgan~\cite{ctgan} and \ctab~\cite{ctabgan} improve data synthesis by introducing several preprocessing steps for categorical, continuous or mixed data types which encode data columns into suitable form for GAN training. The conditional vector designed by \ctgan and later improved by \ctab also helps the GAN training to reduce mode-collapse on minority categories. 
% \ctabplus~\cite{ctabplus}, and \itgan~\cite{itgan} generate tabular data without risking privacy of original data by either adopting differential privacy or controlling the negative log-density of real records during the GANs training. 
Previous research on GANs for tabular data synthesis includes \tablegan~\cite{tablegan}, which enhances column dependencies in synthetic data through the use of an auxiliary classification model in conjunction with discriminator training, and \ctgan~\cite{ctgan} and \ctab~\cite{ctabgan}, which improve data synthesis through preprocessing steps for categorical, continuous, or mixed data types and the use of a conditional vector to reduce mode collapse on minority categories. \ctabplus~\cite{ctabplus} and \itgan~\cite{itgan} also generate tabular data while protecting the privacy of the original data through the use of differential privacy or by controlling the negative log-density of real records during GAN training. 
{While there have been recent works exploring the use of variational autoencoders (VAE)\cite{ctgan}, diffusion models\cite{tabddpm}, and large language models (LLMs)~\cite{great} to synthesize tabular data, GANs remain the most widely used and efficient method in this area. This is what motivates us to incorporate the GAN structure into VFL.} 
% There are also works using variational autoencoder (VAE)~\cite{ctgan}, diffusion model~\cite{tabddpm} and large language models (LLMs)~\cite{great} to synthesize tabular data.  But GAN is still the most efficient and widest used methods in this area, which motivates us to incorporate GAN structure into VFL.
% Even though there are also works using variational autoencoder (VAE)~\cite{ctgan}, diffusion model~\cite{tabddpm} and large language models (LLMs)~\cite{great} to synthesize tabular data.  GANs are still the most efficient and widest used methods in this area.  
% By proposing the privacy-preserving feature encoding and table-similarity aware aggregation weights, Fed-TGAN~\cite{fedtgan} is the first framework that incorporates state-of-the-art tabular GAN into HFL system. To our best knowledge, \algo is the first framework that adapts state-of-the-art tabular GAN into VFL.  
Fed-TGAN~\cite{fedtgan} is the first framework to incorporate a SOTA tabular GAN into a horizontal federated learning system through the use of privacy-preserving feature encoding and table-similarity aware aggregation weights. To the best of our knowledge, \algo is the first framework to adapt a SOTA conditional tabular GAN for vertical federated learning. 
% \cite{aegan} is the first paper to discuss the permutation invariant in table synthesis. It uses the same framework as \itgan, which uses an autoencoder to transform input table into latent vector. Since the
% dimension of the latent vector is much lower than original tabular data. The permutation has less impact on the input representation. Drawback of these methods is also clear that there is loss of information during the transformation from table to latent vector. Therefore, GAN cannot learn the knowledge from the information that loses during this compression.

%  \begin{figure}[t]
%     \centering
%     \includegraphics[width=0.92\linewidth]{figures/D20G02_num_clients_new.png}
%     \vspace{-0.7em}
%     \caption{Under $D^{2}_{0}G^{0}_{2}$ setting, the difference between real and \algo data over varied numbers of clients: default generator (firm line), enlarged generator (dashed line). Corr.Diff. (right scale)}
%     \label{fig:d20g02_num_clients}
%    \vspace{-1.2em}
% \end{figure}
\section{Conclusion}
%In order to meet the demand for generating tabular data in a distributed environment, 
In this paper, we propose a framework \algo which trains tabular data synthesizer via vertical federated learning. \algo designs a privacy-preserving architecture which allows to train state-of-the-art tabular GAN with distributed clients where each of them contains unique features.  \algo also proposes a training mechanism called \textit{training-with-shuffling} in order to adopt the conditional vector into tabular GAN training. Results show that \algo can generate synthetic data that capture the column dependencies from different clients, the synthetic data quality is comparable to the one generated by the centralized tabular GAN, as low as to 2.7\% difference on machine learning utility. \algo can maintain stable on synthetic data column distribution even under extreme imbalanced data distribution across clients and different number of clients. Differential Privacy (DP) is evaluated on \algo to safeguard against Membership Inference Attacks. The results indicate that DP can effectively protect \algo; however, with a significant cost to the utility of synthetic data.
%making it a robust and scalable solution for generating synthetic tabular data in a distributed and privacy-preserving manner. 
Beyond the presented work, one of the challenges is to investigate measures to counteract collusion between the server and clients. Another interesting area of research is how to defend against data poisoning attacks.

\bibliographystyle{ACM-Reference-Format}
\bibliography{sample}

\end{document}